\pdfoutput=1 
\documentclass{article}
\usepackage{aaai25}  
\usepackage{times}  
\usepackage{helvet}  
\usepackage{courier}  
\usepackage{graphicx} 
\usepackage{natbib}  
\usepackage{caption} 
\usepackage{url}            
\usepackage{booktabs}       
\usepackage{amsfonts}       
\usepackage{nicefrac}       
\usepackage{microtype}      
\usepackage{xcolor}         

\usepackage{soul}

\usepackage{afterpage}
\usepackage{graphicx}
\usepackage{makeidx}
\usepackage{amsmath,amssymb,amsfonts,amsthm,mathtools}
\usepackage{algpseudocode}
\usepackage{textcomp}
\usepackage{xcolor}
\usepackage{array}
\usepackage{times}
\usepackage{url}
\usepackage{caption}

\usepackage{booktabs}
\usepackage[linesnumbered,ruled,vlined]{algorithm2e}
\usepackage{multirow}
\usepackage{color}
\usepackage{comment}
\usepackage{graphics}
\usepackage{rotating}
\usepackage{epstopdf}
\usepackage{lscape}
\usepackage{thm-restate}
\usepackage{amsmath}
\usepackage{bm}
\usepackage{amssymb}
\usepackage{amsthm}
\usepackage{mathtools}
\usepackage{multirow}
\usepackage{multicol}
\usepackage{subcaption}
\usepackage{tabularx}
\usepackage{xcolor}
\usepackage{listings}
\usepackage{chngcntr}
\usepackage{amsfonts}
\usepackage{blkarray}
\usepackage[capitalise]{cleveref}
\usepackage{adjustbox}

\newtheorem{theorem}{Theorem}

\newtheorem{lemma}[theorem]{Lemma}

\theoremstyle{definition}
\newtheorem{definition}[theorem]{Definition}

\theoremstyle{remark}
\newtheorem{remark}[theorem]{Remark}

\newcommand{\sd}[1]{{{\footnotesize±}{\scriptsize#1}}}

\newcommand{\calD}{\mathcal{D}}
\newcommand{\calE}{\mathcal{E}}

\newcommand{\calL}{\mathcal{L}}

\newcommand{\calV}{\mathcal{V}}

\newcommand{\bbH}{\mathbf{H}}

\DeclareCaptionStyle{ruled}{labelfont=normalfont,labelsep=colon,strut=off} 
\title{Asymmetric Learning for Spectral Graph Neural Networks}
\author{
    Fangbing Liu,\textsuperscript{\rm 1}
    Qing Wang\textsuperscript{\rm 1} \\
}
\affiliations {
    \textsuperscript{\rm 1}Graph Research Lab, School of Computing, Australian National University\\
    fangbing.liu@anu.edu.au, qing.wang@anu.edu.au
}

\usepackage{bibentry}

\begin{document}

\maketitle

\begin{abstract}

Optimizing spectral graph neural networks (GNNs) remains a critical challenge in the field, yet the underlying processes are not well understood. In this paper, we  investigate the inherent differences between graph convolution parameters and feature transformation parameters in spectral GNNs and their impact on the optimization landscape. Our analysis reveals that these differences contribute to a poorly conditioned problem, resulting in suboptimal performance. To address this issue, we introduce the concept of the block condition number of the Hessian matrix, which characterizes the difficulty of poorly conditioned problems in spectral GNN optimization. We then propose an asymmetric learning approach, dynamically preconditioning gradients during training to alleviate poorly conditioned problems. Theoretically, we demonstrate that asymmetric learning can reduce block condition numbers, facilitating easier optimization. Extensive experiments on eighteen benchmark datasets show that asymmetric learning consistently improves the performance of spectral GNNs for both heterophilic and homophilic graphs. This improvement is especially notable for heterophilic graphs, where the optimization process is generally more complex than for homophilic graphs. Code is
available at https://github.com/Mia-321/asym-opt.git.
\end{abstract}

\section{Introduction}


Graph Neural Networks (GNNs) have been extensively studied in recent years~\citep{gnn_survey,loukas2020graph,dwivedi2020benchmarking,gnn_att,gin}.
Inspired by the seminal work of \citep{gcn,chebynet},
research has delved into spectral GNNs, which extract information from graph-structured data by performing graph convolutions in the spectral domain~\citep{wang2022powerful,analy_express,yang2022new}. Generally, a \emph{spectral GNN} is  defined as 
\begin{equation}\label{eq_def_spectral_gnn}
\hat{Y} = \sigma(g_{\Theta}(M)f_W(X)).
\end{equation}
Here, $M\in \mathbb{R}^{n\times n}$ is a graph matrix (e.g., normalized Laplacian), $X\in \mathbb{R}^{n\times d}$ is a feature matrix,
$g_{\Theta}$ is a {graph convolution function} parameterized by $\Theta$, $f_{W}$ is a {feature transformation function} parameterized by $W$,
and $\sigma$ is a non-linear function (e.g., softmax). For computational efficiency and local vertex analysis~\citep{wavelets}, the graph convolution function is typically approximated by polynomial filters in the form $g_{\Theta}(M)= \sum_{k=0}^K \theta_k T_k(M)$,
where each $T_k(\bullet)$ denotes the $k$-th polynomial basis. The exact form of $T_k$ is determined by the selected polynomial basis, e.g., the Chebyshev basis~\citep{chebynet,chebii} or the Bernstein basis~\citep{bernnet}.



\begin{figure}[t!]
    \centering
    \begin{subfigure}[b]{0.12\textwidth}
        \centering
        \includegraphics[width=\textwidth]{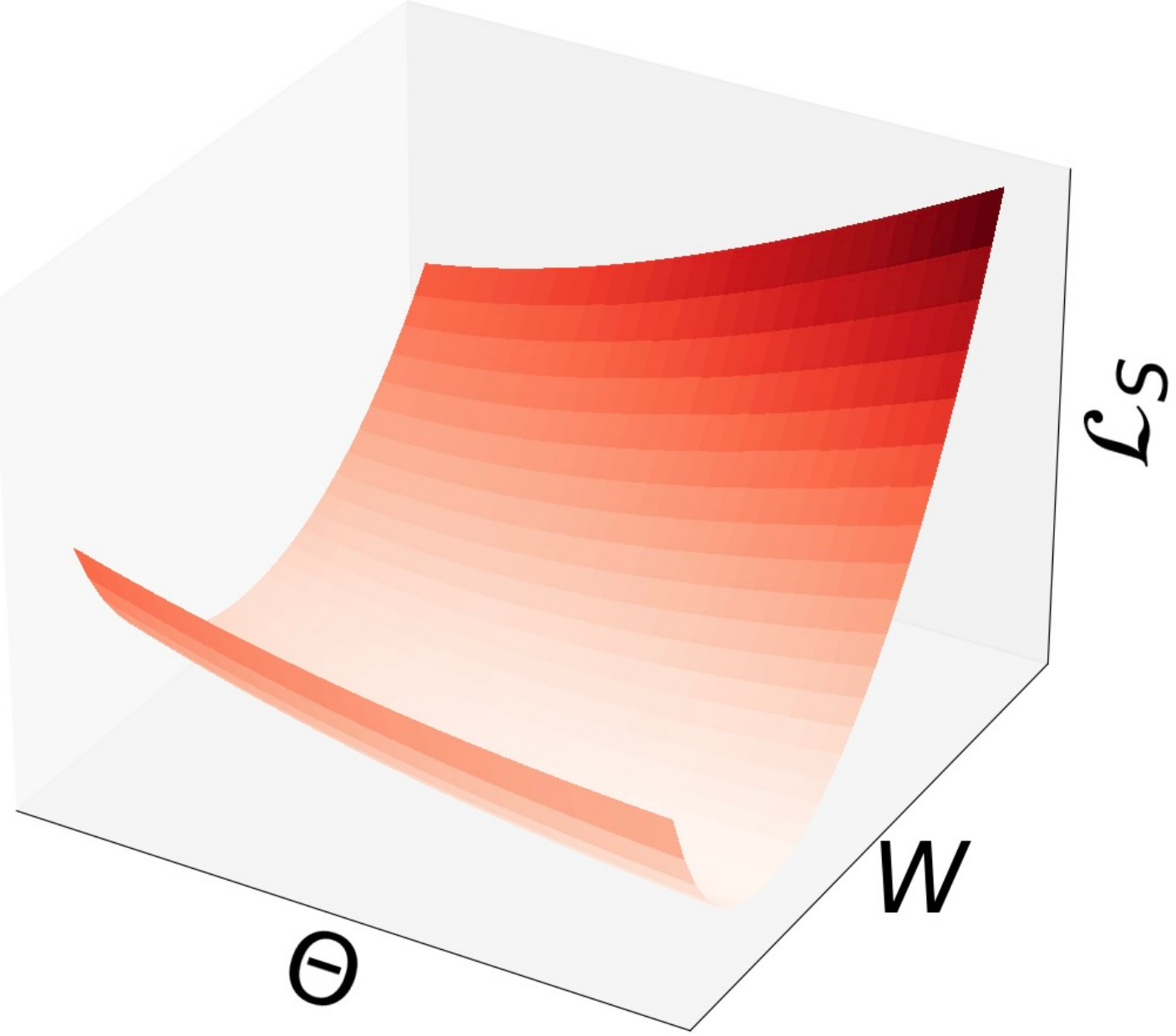}
        \caption{Texas-50}
        \label{fig_iter_surface_texas-50}
    \end{subfigure} \hspace{1cm}
    \begin{subfigure}[b]{0.12\textwidth}
        \centering
        \includegraphics[width=\textwidth]{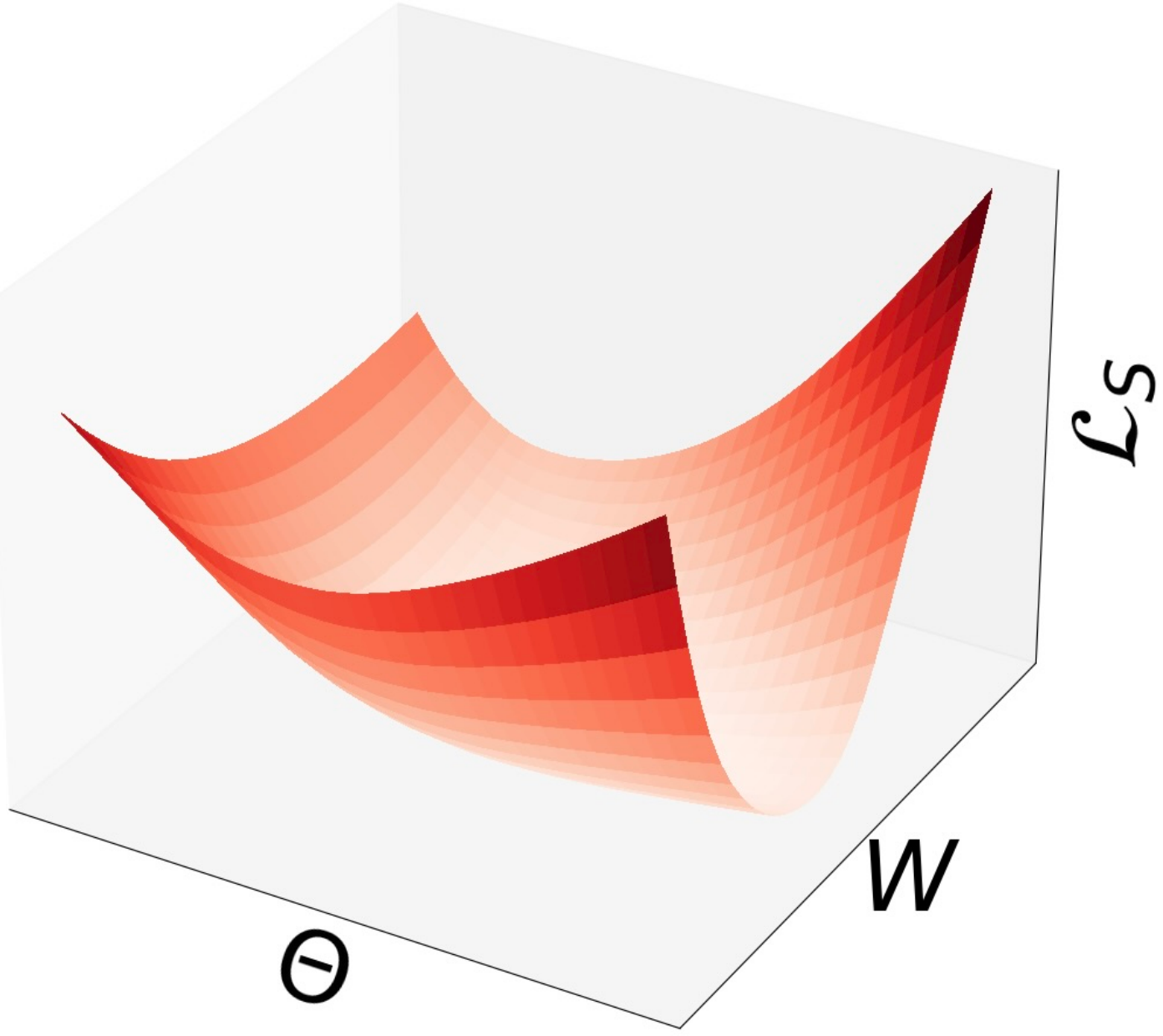}
        \caption{Texas-200}
        \label{fig_iter_surface_texas-200}
    \end{subfigure}
    \vskip\baselineskip
    \begin{subfigure}[b]{0.12\textwidth}
        \centering
        \includegraphics[width=\textwidth]{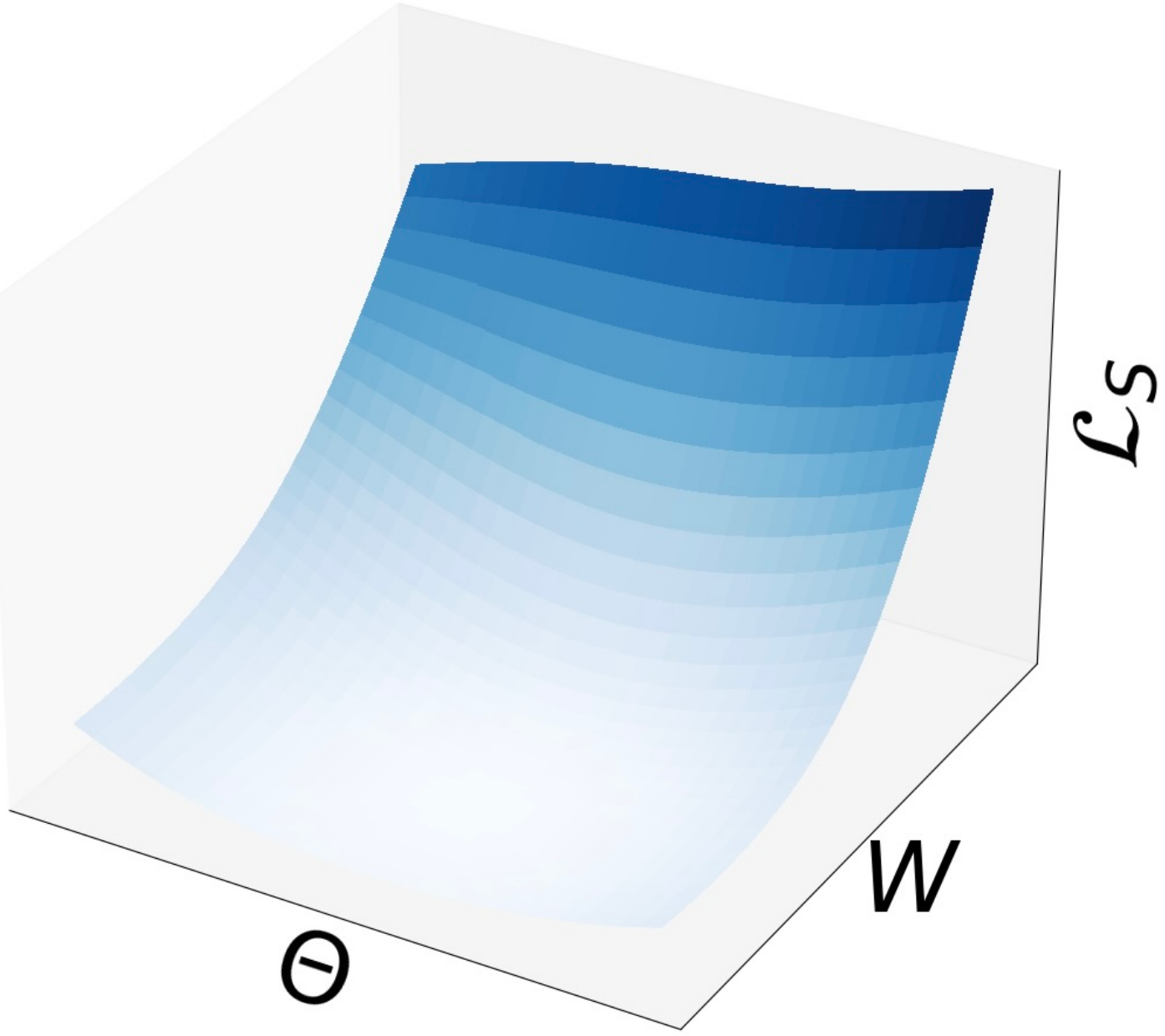}
        \caption{Cora-50}
        \label{fig_iter_surface_cora-50}
    \end{subfigure}\hspace{1cm}
    \begin{subfigure}[b]{0.12\textwidth}
        \centering
        \includegraphics[width=\textwidth]{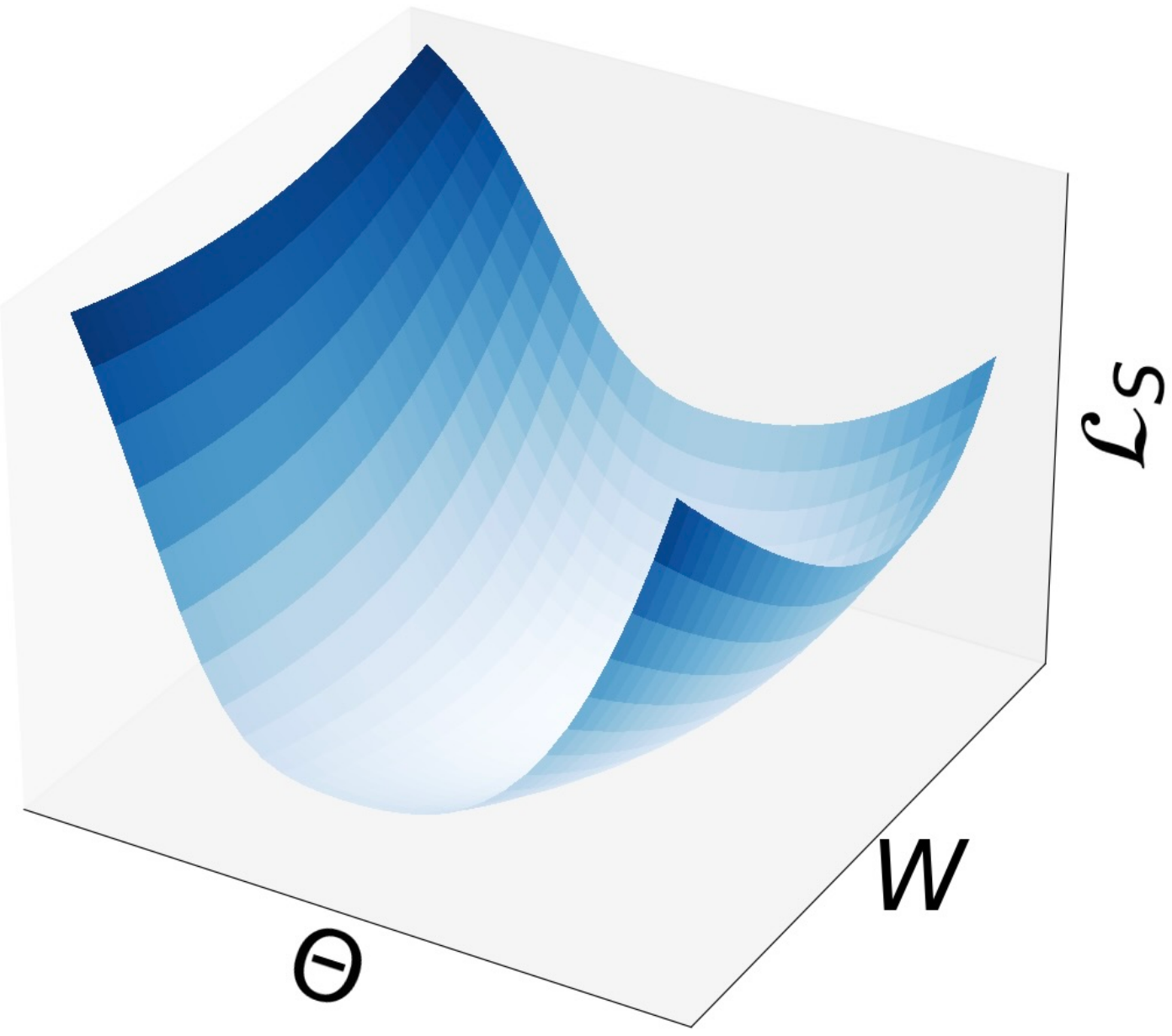}
        \caption{Cora-200}
        \label{fig_iter_surface_cora-200}
    \end{subfigure}
    \caption{The landscapes of ChebNet on Texas and Cora at iterations $50$ and $200$. The $x,y,z$-axes represent the directions of $\Theta,W$, and the empirical loss $\calL_S$, respectively. The landscapes change during training, and the stretched direction of the parameters $\Theta$ and $W$ also shifts.}
    \label{fig_asym_iter_surface}
\end{figure}

Despite the success of spectral GNNs in graph learning tasks ~\citep{beyondh,linkx,GeomGCNGG},  their training process remains inadequately understood.
Notably, the parameters $\Theta$ and $W$ play distinct roles in modeling and may significantly differ in dimensions. Specifically, the dimension of $\Theta$ is fewer than $20$ while $W$ has thousands of dimensions. During the training of spectral GNNs, different learning rates for parameters $\Theta$ and $W$ are often adopted to achieve good performance. Interestingly, we observe that the contours of the loss function undergo stretching or compression along specific directions during training. In stretched directions, small alterations in parameters lead to significant changes in the loss function. Figure~\ref{fig_asym_iter_surface} shows the loss landscapes of a spectral GNN model, ChebNet~\citep{chebynet}. There always exists a stretched direction for $\Theta$ or $W$ during training, indicating that parameters $\Theta$ and $W$ exhibit different sensitivities to learning rates and gradient updates.



\begin{figure}[t!]
    \centering
    \includegraphics[width=0.8\columnwidth ]{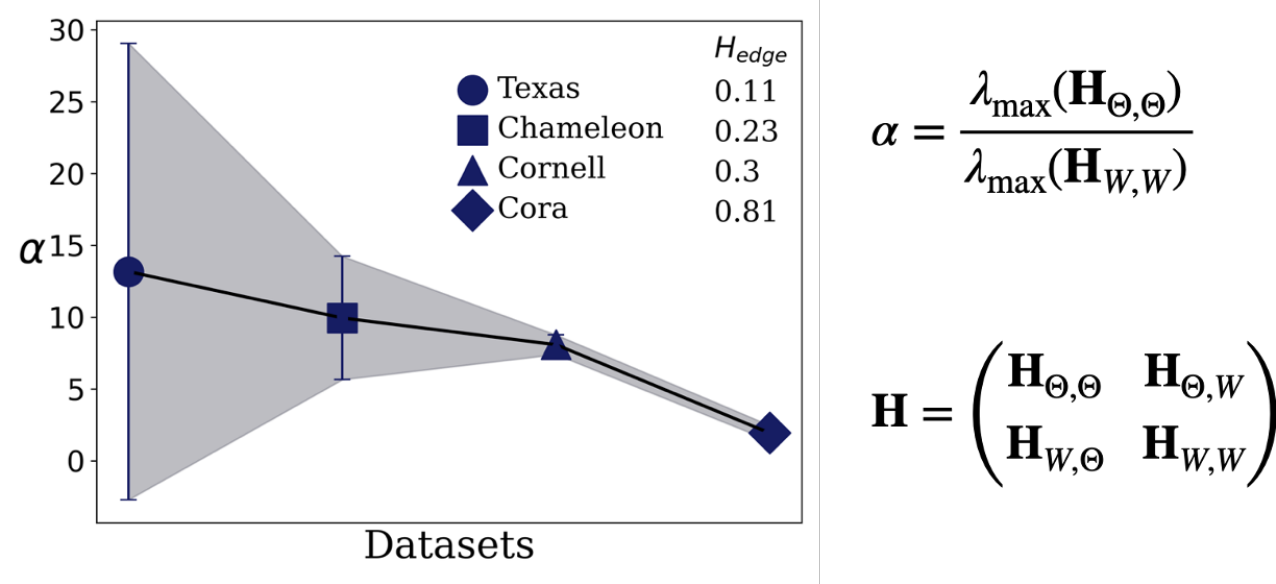}
    \caption{The largest eigenvalue ratio $\alpha$ of ChebNet when applied to graphs of varying edge homophily ratios $H_{edge}$~\citep{beyondh} for node classification. $\bbH$ represents the Hessian matrix and variances are depicted as grey shades. \vspace*{-0cm}
    }   
\label{fig_asym_eigenvalue_gap}
\end{figure}

The distorted landscapes of spectral GNNs can be further explained from the view of Hessian matrix.
According to~\citep{rate_eigenvalue-1,rate_eigenvalue-2}, neural networks exhibits non-convex characteristics and the optimal learning rate of a parameter is inversely proportional to the largest eigenvalue of the Hessian matrix. The Hessian matrix $\bbH$ of a spectral GNN contains second partial derivatives of $\Theta$ and $W$, which can be divided into two diagonal block matrices $\bbH_{\Theta,\Theta}$ and $\bbH_{W,W}$, and two off-diagonal block matrices. In scenarios where $\Theta$ and $W$ exhibit minimal interference, the optimal learning rates for $\Theta$ and $W$ correspond inversely to the largest eigenvalues of their respective diagonal block Hessian matrices. 
The largest eigenvalue of a Hessian matrix indicates the maximum rate of change of the gradient with respect to a small change of its corresponding parameter. In regions where the largest eigenvalue of a Hessian matrix is large, the gradient norm tends to be large as well, indicting a large loss change with respect to a small parameter change~\citep{book-numerical}. The larger the gap between two diagonal block matrices of Hessian with respect to $\Theta$ and $W$ respectively, the more distorted of the local landscape.
Figure~\ref{fig_asym_eigenvalue_gap} shows the ratio of the largest eigenvalues of two diagonal block Hessian matrices (largest eigenvalue ratio) when a spectral GNN is applied on different graphs. There is a notable disparity in the largest eigenvalue ratio across different graphs. 
When a graph is more homophilic, the largest eigenvalue ratio is closer to one, indicting a less distorted landscape. For example, the landscapes of Texas are more distorted than those of Cora because Texas is a heterophilic graph while Cora is a homophilic graph. 

To characterize the aforementioned phenomenon, particularly the relationship between the parameters $\Theta$ and $W$ in spectral GNNs and their impact on optimization, we introduce the concept of the \emph{block condition number} of the Hessian matrix. This metric serves as an indicator for assessing the difficulty of an optimization problem. Specifically, utilizing the block condition number, we show that poorly conditioned spectral GNNs are characterized by a large block condition number of the Hessian matrix.

A widely adopted strategy for addressing poorly conditioned optimization problems is \emph{preconditioning}~\citep{Saad2003iterative}. 
Advanced adaptive learning rate optimizers
tackle poorly conditioned problems by assigning different learning rates to different parameters, calculated based on a gradient history~\citep{adam,AdaGrad}. However, these optimizers overlook the inherent parameter magnitude differences exhibited by the parameters $\Theta$ and $W$ during the optimization of spectral GNNs. To accelerate the learning process and potentially achieve superior solutions, we propose \emph{asymmetric learning} for training spectral GNNs. This new method involves scaling the gradient of the empirical loss function asymmetrically during each iteration. The scaling factors within the asymmetric preconditioner are determined by the \emph{gradient-parameter norm ratio},  
which is closely associated with the largest eigenvalues of the block Hessian matrices $\bbH_{\Theta,\Theta}$ and $\bbH_{W,W}$. Notably, this asymmetric preconditioning of the gradient corresponds to an effective preconditioner for the Hessian matrix $\bbH$. 
Under reasonable assumptions, we demonstrate that asymmetric preconditioning can effectively reduce the block condition number of the Hessian matrix $\bbH$, thereby facilitating more efficient optimization of spectral GNNs.


\paragraph{Contributions.}
In summary, our contributions are: (1) We reveal that the inherent difference between the graph convolution parameter $\Theta$ and the feature transformation parameter $W$ in spectral GNNs leads to a poorly conditioned optimization problem.
(2) We introduce the concept of block condition number to quantify optimization difficulty.
(3) We introduce an asymmetric learning approach to improve the training and performance of spectral GNNs.
(4) Our extensive experiments on eighteen benchmark datasets demonstrate that asymmetric learning consistently enhances spectral GNN performance across various graphs.

\section{Problem Analysis}~\label{sect_prob_analysis}
Let $G=\{\calV,\calE \}$ be an undirected graph with a node set $\calV$, an edge set $\calE$, and $n=|\calV|$. Each node $v_i$ is associated with a feature vector $x_i$ and a label $y_i$. We use $X$ to denote the node feature matrix, where the $i$-th row corresponds to $x_i$. Let $S=(X,\{y_i\}_{i=1}^m)$ be the training set containing $m$ labeled nodes randomly selected from all nodes of the graph. Similarly, we obtain the validation set $\calD_{val}=(X,\{y_i\}_{i=m+1}^{m+q})$ containing $q$ labeled nodes. 
A loss function $\ell(\hat{y}_i,y_i;\Theta,W)$ estimates the distance between the truth node label $y_i$ and the prediction $\hat{y}_i$ from a spectral GNN parameterized by $\Theta,W$ in \cref{eq_def_spectral_gnn}. The empirical loss is $\calL_S(\Theta,W)=\frac{1}{m} \sum_{i=0}^m \ell(\hat{y}_i,y_i;\Theta,W)$, and the validation loss is $\calL_{\calD_{val}}(\Theta,W)=\frac{1}{q} \sum_{i=m+1}^{m+q} \ell(\hat{y}_i,y_i;\Theta,W) $. We use $||\bullet||_2$ to denote the $\ell_2$ norm of $\bullet$, $vec(\bullet)$ to vectorize a matrix by concatenating its rows, and
$d_{\Theta}=|\Theta|$ and $,d_W=|vec(W)|$ to denote the total parameter number of $\Theta$ and $W$, respectively. 

\paragraph{Poorly Conditioned Spectral GNNs.}
We begin by presenting the classic definition of a poorly conditioned problem and then introduce a new concept called the \emph{block condition number}. Using this new concept, we analyze the poorly conditioned nature of spectral GNNs.

\begin{definition}[Poorly Conditioned Problem~\citep{matrix_computation}]
Let $\calL_S(\Psi)$ be a convex objective function, parameterized by $\Psi$, and $\bbH$ be its associated Hessian matrix. If $\bbH$ is ill-conditioned, the optimization of $\calL_S(\Psi)$ is poorly conditioned.  The extent to which $\bbH$ is ill-conditioned is quantified by the \emph{condition number} of $\bbH$, defined as
\begin{equation}
    \kappa(\bbH) = \frac{\lambda_{max}(\bbH)}{\lambda_{min}(\bbH)}.
\end{equation}
Here, $\kappa(\bbH)$ is the ratio of the absolute values of the maximum eigenvalue $\lambda_{max}(\bbH)$ to the minimum eigenvalue $\lambda_{min}(\bbH)$. 
\end{definition}

\smallskip
\begin{remark}
Eigenvectors corresponding to the maximum eigenvalue $\lambda_{max}(\bbH)$ and the minimum eigenvalue $\lambda_{min}(\bbH)$ represent the most and least sensitive directions of the loss function $\calL_S$, respectively. A significant gap between $\lambda_{max}(\bbH)$ and $\lambda_{min}(\bbH)$ induces numerical instability during optimization and slow down the convergence rate~\citep{book-numerical}.
\end{remark}

In spectral GNNs, the Hessian matrix $\bbH$ can be partitioned into two diagonal blocks $\bbH_{\Theta,\Theta}$ and $\bbH_{W,W}$, and two non-diagonal blocks $\bbH_{\Theta,W}$ and $\bbH_{W,\Theta}$. The $ij$-th element of $\bbH_{\Theta,\Theta}$ is $\frac{\partial^2 \calL_S(\Theta,W)}{\partial \Theta_i \partial \Theta_j}$, where $i,j \in [1,d_{\Theta}]$. The $ij$-th element of $\bbH_{W,W}$ is $\frac{\partial^2 \calL_S(\Theta,W)}{\partial W_i \partial W_j}$, where $i,j \in [1,d_W]$. The $ij$-th element of $\bbH_{\Theta,W}$ (equivalently $\bbH_{W,\Theta}^T$) is  $\frac{\partial^2 \calL_S(\Theta,W)}{\partial \Theta_i \partial W_j}$, with $i \in [1,d_{\Theta}]$ and $j \in [1,d_W]$.


To understand how the parameters $\Theta$ and $W$ influence the optimization process, we propose the notion of block condition number for spectral GNNs.

\begin{definition}[Block Condition Number]
For a spectral GNN parameterized by \(\Theta\) and \(W\), the \emph{block condition number} of its Hessian matrix \(\bbH\) with respect to the empirical loss is defined as:
\begin{equation}
    \kappa'(\bbH)= \frac{\max(\lambda_{\max}(\bbH_{\Theta,\Theta}),\lambda_{\max}(\bbH_{W,W}) )}{\min(\lambda_{\max}(\bbH_{\Theta,\Theta}),\lambda_{\max}(\bbH_{W,W}) ) }.
\end{equation}

\end{definition}

\smallskip

\begin{remark}
The block condition number $\kappa'(\bbH)$ serves as a metric to quantify the disparity between the maximum eigenvalues of the block Hessian matrices \(\bbH_{\Theta,\Theta}\) and \(\bbH_{W,W}\). A spectral GNN is considered \emph{poorly conditioned} if \(\kappa'(\bbH)\) is large.
According to the interlacing theorem~\citep{matrix_analysis}, when a symmetric matrix is divided into blocks according to parameters, the eigenvalues of the submatrices interlace with the eigenvalues of the full matrix. In the case where the Hessian matrix $\bbH$ is positive semi-definite, we have $\max(\lambda_{max}(\bbH_{\Theta,\Theta}),\lambda_{max}(\bbH_{W,W}) )\leq \lambda_{max}(\bbH)$ and $\lambda_{min}(\bbH)\leq \min(\lambda_{max}(\bbH_{\Theta,\Theta}),\lambda_{max}(\bbH_{W,W}))$. Therefore, $\kappa'(\bbH)\leq \kappa(\bbH)$. A problem identified as poorly conditioned using $\kappa'$ will also be poorly conditioned when assessed using $\kappa$. However, the converse does not hold true. Thus, the block condition number offers a more stringent criterion for determining whether a problem is poorly conditioned, as it directly models the impact of the differences between the parameters $\Theta$ and $W$ on the optimization process. 
\end{remark}


In spectral GNNs, the block condition number for heterophilic graphs tends to be larger than that for homophilic graphs. When $\lambda_{max}(\bbH_{\Theta,\Theta})>\lambda_{max}(\bbH_{W,W})$, the block condition number $\kappa'(\bbH)$ is reduced to the largest eigenvalue ratio shown in Figure~\ref{fig_asym_eigenvalue_gap}, i.e., $\kappa'(\bbH)=\alpha$ when applying ChbNet on those four datasets. Thus, the optimization on heterophilic graphs is generally more difficult than the optimization on homophilic graphs.


The relation between the optimal learning rate $\eta^*$ and the Hessian matrix $\bbH$ is that $\eta^* = \frac{1}{\lambda_{max}(\bbH)}$~\citep{rate_eigenvalue-2}. For spectral GNNs, we observe that $\lambda_{max}(\bbH_{\Theta,\Theta}) \leq \lambda_{max}(\bbH)$ and $\lambda_{max}(\bbH_{W,W}) \leq \lambda_{max}(\bbH)$. Consequently, the learning rates must satisfy $\eta \leq \eta_{\Theta}$ and $\eta \leq \eta_W$, where $\eta$, $\eta_{\Theta}$, and $\eta_W$ denote the learning rates for all parameters, $\Theta$, and $W$, respectively. Given that $\lambda_{max}(\bbH_{\Theta,\Theta}) \neq \lambda_{max}(\bbH_{W,W})$ in most cases, employing different learning rates for $\Theta$ and $W$ is a common practice in the training of spectral GNNs, as evidenced in recent studies~\citep{bernnet,chebii}.

\section{Asymmetric Learning}\label{sect_asym_learning}


In this section, we first introduce the \emph{gradient-parameter norm ratio} (GPNR), which is computed based on the magnitudes of gradients and parameter values. We then discuss the relationship between GPNR and the maximum eigenvalue of the Hessian matrix. After that, we propose \emph{asymmetric preconditioning} on the gradients of spectral GNNs and examine its impact on the Hessian matrix. We theoretically demonstrate that our asymmetric preconditioning can mitigate the poorly conditioned optimization of spectral GNNs under mild assumptions.

\begin{figure}[htbp!]
    \centering
    \includegraphics[width=0.35\textwidth]{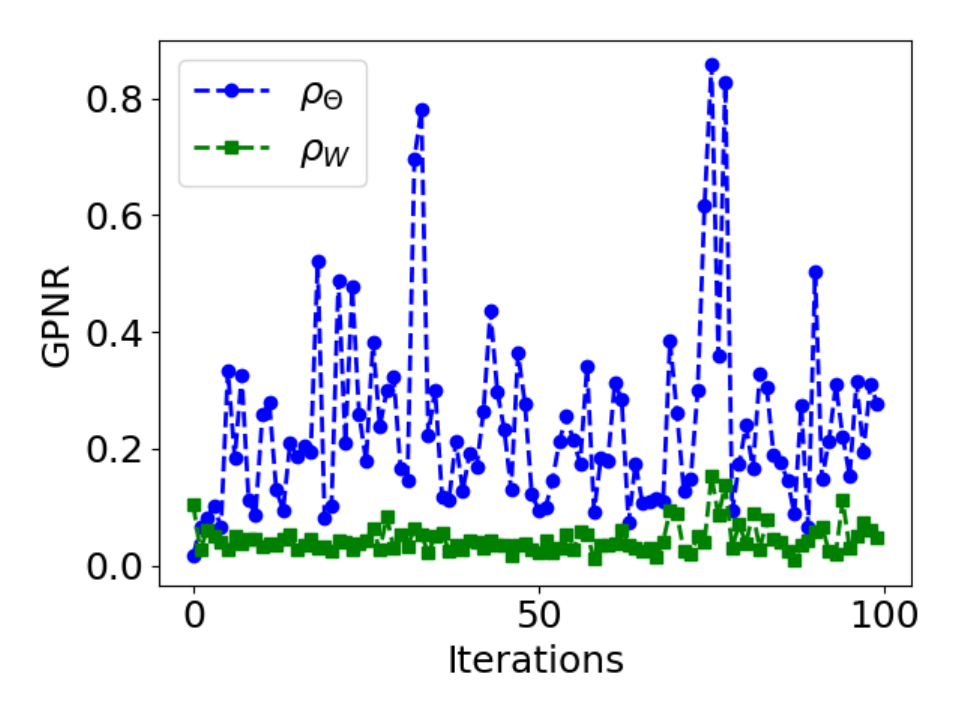}
    \caption{The Gradient-Parameter Norm Ratio (GPNR) of $\rho_{\Theta}$ is larger than $\rho_W$ during iterations. It means the updating speed of $\Theta$ is faster than that of $W$.
 }
    \label{fig_upd_speed}
\end{figure}

\paragraph{Gradient-Parameter Norm Ratio}
The update rule for gradient descent in spectral GNNs is given by
\begin{align}
    \begin{split}
        \Theta^t &= \Theta^{t-1} - \eta \nabla_{\Theta} \calL_S(\Theta^t,W^t);\\
        \quad W^t &= W^{t-1} - \eta \nabla_W \calL_S(\Theta^t,W^t),
    \end{split}
\end{align}
where $\eta$ denotes the learning rate.

We introduce the notion of the gradient-parameter norm ratio  to estimate the update speed of parameters.
\begin{definition}[Gradient-Parameter Norm Ratio]
For a differentiable function $\calL_S$ parameterized by $\Psi$, the gradient-parameter norm ratio (GPNR) of $\Psi$ is defined as
\begin{equation}
    \rho_{\Psi}=\frac{||\nabla \calL_S(\Psi) ||_2 }{||\Psi ||_2 },
\end{equation}
where $\nabla \calL_S(\Psi)$ is the gradient of $\calL_S$ with respect to $\Psi$.
\end{definition}

\begin{remark}
The GPNRs of parameters $\Theta^t$ and $W^t$ are $\rho_{\Theta}^t = \frac{||\nabla_{\Theta}\calL_S(\Theta^t,W^t)||_2}{||\Theta||_2}$ and $\rho_W^t = \frac{||\nabla_W\calL_S(\Theta^t,W^t)||_2}{||W||_2}$, respectively, at each iteration $t$. These ratios are effective metrics quantifying the update speeds of $\Theta$ and $W$. 
Generally, parameters should be updated at similar speeds~\citep{lars}, meaning that parameters with larger magnitudes undergo larger updates. 
However, our analysis of $\rho_{\Theta}^t$ and $\rho_W^t$, as shown in Figure~\ref{fig_upd_speed}, reveals that $\rho_{\Theta}^t$ is typically larger than $\rho_W^t$, indicating that $\Theta$ is updated more rapidly than $W$.
\end{remark}




We further demonstrate the close relationship between the GPNR and the Hessian matrix. 

\begin{restatable}[Point Proximity]{assumption}{assumpsmalldiffnorm}\label{assumption-asym-small-diff-norm}
Let $\Psi$ be a point near a critical point $\Psi^*$. If \(\|\Psi - \Psi^*\|_2\) is close to zero, 
we assume $\|\Psi - \Psi^*\|_2\leq \|\Psi\|_2$.
\end{restatable}
\begin{remark}
    In spectral GNNs, $\Psi=\left[ \Theta;W \right]$. As the dimension of $W$ is usually in thousands, there is a high probability that $\|\Psi\|_2$ is greater than zero. Thus, when \(\|\Psi - \Psi^*\|_2\) is a very small number approximating zero, it is natural to assume that $\|\Psi - \Psi^*\|_2 \leq \|\Psi\|_2$. 
\end{remark}

\begin{restatable}[GPNR and Maximum Eigenvalue of Hessian]{proposition}{propositionasymgpnr}\label{proposition-asym-gpnr}
For a neural network parameterized by $\Psi$ and the Hessian matrix $\bbH_{\Psi}$ of empirical loss at point $\Psi$,
when $\Psi$ is near a critical point $\Psi^*$ under~\cref{assumption-asym-small-diff-norm}, we have $\rho_{\Psi} \leq \lambda_{max}(\bbH_{\Psi})$.
\end{restatable}

\begin{remark}
The Hessian matrix $\bbH_{\Psi}$ provides insights into how  gradients change in the vicinity of the point $\Psi$. The empirical loss $\calL_S$ changes rapidly along the director of the eigenvector corresponding to $\lambda_{max}(\bbH_{\Psi})$.
Consequently, a small perturbation around $\Psi$ can result in a significant increase in the gradient $\nabla_{\Psi} \calL_S(\Psi)$, leading to a large $\rho_{\Psi} = \frac{||\nabla_{\Psi} \calL_S(\Psi)||_2}{||\Psi||_2}$. When the parameter is near a critical point, $\lambda_{max}(\bbH_{\Psi})$ 
is greater than or equal to the GPNR. 
Hence, if the GPNR is large, then $\lambda_{max}(\bbH_{\Psi})$ is also large.
\end{remark}

\paragraph{Asymmetric Preconditioner}
We now define asymmetric
preconditioning on the gradients of spectral GNNs. Let
{\small
\begin{align}\label{eq_asym_grad_scale}
\begin{split}
    s_{\Theta}^t = \frac{||\Theta^t||_2}{||\nabla_{\Theta}\calL_S(\Theta^t,W^t)||_2}; \quad
    s_W^t = \frac{||W^t||_2}{||\nabla_W \calL_S(\Theta^t,W^t)||_2}.
\end{split}
\end{align}
}

Then, the \emph{asymmetric preconditioner} for $s_{\Theta}^t$ and $s_W^t$ is a diagonal preconditioner, defined as
  \begin{equation}\label{eq_asym_diag_conditioner}
    R^t = \begin{pmatrix}
            s_{\Theta}^t\mathbf{I_{d_{\Theta}}} & \mathbf{0}\\
            \mathbf{0} &
            s_W^t\mathbf{I_{d_W}}
        \end{pmatrix}
\end{equation}
where $\mathbf{I_{d_{\Theta}}}$ and $ \mathbf{I_{d_W}}$ are the identity matrices of dimensions $d_{\Theta}$ and $d_W$, respectively. This yields new gradient updates
\begin{align}\label{eq_asym_new_grad}
    \begin{split}
        &[\overline{\nabla}_{\Theta}\calL_S(\Theta^t,W^t); \overline{\nabla}_W \calL_S(\Theta^t,W^t)] \\
        &\hspace{1.5cm}= R^t [\nabla_{\Theta}\calL_S(\Theta^t,W^t); \nabla_W \calL_S(\Theta^t,W^t)].
    \end{split}
\end{align}

The GPNRs are also updated: \( \rho_{\Theta}^t = \frac{||\overline{\nabla}_{\Theta}\calL_S(\Theta^t,W^t)||_2}{||\Theta^t||_2} = 1 \) and \( \rho_W^t = \frac{||\overline{\nabla}_W \calL_S(\Theta^t,W^t)||_2}{||W^t||_2} = 1 \). Consequently, the parameters \( \Theta \) and \( W \) are updated at the same rate, preventing overly aggressive or insufficient updates.

Asymmetric preconditioning affects the Hessian matrix, thereby altering the optimization landscape and the block condition number. The new Hessian matrix is given by
\begin{equation}\label{eq_asym_pre_hesian}
    \bbH'= R^t \bbH.
\end{equation} 
We will prove later that the block condition number decreases after preconditioning, i.e., $\kappa'(\bbH')\leq \kappa'(\bbH)$.





\paragraph{Moving Average of Parameter Norms}
To ensure smooth changes in the parameter norms during iterations, especially when an optimizer updates parameters aggressively, we employ the exponential moving average technique~\citep{ema-book}.
The parameter norms \(\pi_{\Theta}^t\) and \(\pi_W^t\) at the \(t\)-th iteration are calculated using the exponential moving average as follows:
\begin{align}\label{eq_para_norm}
\begin{split}
    \pi_{\Theta}^t &= \beta_{\pi_{\Theta}} \pi_{\Theta}^{t-1} + (1-\beta_{\pi_{\Theta}}) ||\Theta^t||_2;\\
    \pi_W^t &= \beta_{\pi_W} \pi_W^{t-1} + (1-\beta_{\pi_W}) ||W^t||_2,
\end{split}
\end{align}
where \(\beta_{\pi_{\Theta}}\) and \(\beta_{\pi_W}\) are the smoothing factors in $[0,1]$ for the parameter norms \(\pi_{\Theta}\) and \(\pi_W\), respectively.



Asymmetric training can be integrated with any optimizer, and the optimization procedure is elaborated in Algorithm~\ref{algorithm_asym}. 
Initially, the weights $\Theta^0$ and $W^0$ are randomly initialized, and the best validation loss $\calL_{\calD_{val}}^b$ is set to infinity. During each iteration, the raw gradient $[\nabla_{\Theta}\calL_S(\Theta^t,W^t); \nabla_W \calL_S(\Theta^t,W^t)]$ of the empirical loss is computed.
Based on the specific update rules of the chosen optimizer $\emph{OP}$, the update vectors $\delta_{\Theta}^t$ and $\delta_W^t$ for parameters $\Theta$ and $W$ are calculated.
Subsequently, the moving average of parameter norms $\pi_{\Theta}^t,\pi_W^t$ and elements $s_{\Theta}^t,s_W^t$ in the asymmetric preconditioner $R^t$ are calculated as follows: 
{
\begin{equation}\label{eq_asym_scale_adam}
    s_{\Theta}^t = \frac{\|\pi_{\Theta}^t \|_2}{\|\delta_{\Theta}^t\|_2}; \quad
    s_W^t = \frac{\|\pi_W^t \|_2}{\|\delta_W^t\|_2}.
\end{equation}
}
The minimum validation loss $\calL_{\calD_{val}}^b$ is then updated, along with the corresponding best parameters $\Theta^b,W^b$.
The new gradient $[\overline{\nabla}_{\Theta}\calL_S(\Theta^t,W^t); \overline{\nabla}_W \calL_S(\Theta^t,W^t)]$ is computed by preconditioning the raw gradient $[\nabla_{\Theta}\calL_S(\Theta^t,W^t); \nabla_W \calL_S(\Theta^t,W^t)]$ using the asymmetric preconditioner $R^t$. This new gradient serves as the input for any optimizer $\emph{OP}$, resulting in new update vectors $\overline{\delta}_{\Theta}^t, \overline{\delta}_W^t$. Consequently, parameters $\Theta^{t+1}$ and $W^{t+1}$ for the next iteration are updated accordingly.

\begin{algorithm}[t!]
\resizebox{0.95\columnwidth}{!}{
\begin{minipage}{\columnwidth}
\caption{Asymmetric Training}
\label{algorithm_asym}
\SetAlgoLined
\SetKwInOut{Input}{Input}
\SetKwInOut{Output}{Output}

\Input{Training set $S=(X,\{y_i\}_{i=1}^m)$, validation set  $\calD_{val}=(X,\{y_i\}_{i=m+1}^{m+q})$, loss function $\ell$, learning rate $\eta$, maximum iteration number $t_{\max}$, and optimizer $\emph{OP}(\cdot$)}
\Output{Trained model}

Initialize weights  $\Theta^0$ and $W^0$, iteration number $t=0$, best validation loss  $\calL_{\calD_{val}}^b = +\infty$\;
\While{$t \leq t_{\max}$}{
  Compute {
  \small $[\nabla_{\Theta}\calL_S(\Theta^t,W^t); \nabla_W \calL_S(\Theta^t,W^t)]$}\;
   {\small $\delta_{\Theta}^t, \delta_W^t \leftarrow \emph{OP}([\nabla_{\Theta}\calL_S(\Theta^t,W^t);\nabla_W \calL_S(\Theta^t,W^t)])$}\; 
    Set parameter norms $\pi_{\Theta}^t$ and $\pi_W^t$ by~\cref{eq_para_norm}\;
  Compute $s_{\Theta}^t,s_W^t$ by~\cref{eq_asym_scale_adam}\; 
  Compute asymmetric preconditioner $R^t$ by~\cref{eq_asym_diag_conditioner}\;
  Compute validation loss $\calL_{\calD_{val}}(\Theta^t,W^t)$\;
  \If{$\calL_{\calD_{val}}(\Theta^t,W^t) \leq \calL_{\calD_{val}}^b$}{
    $\Theta^b = \Theta^t$, $W^b = W^t$\;
$\calL_{\calD_{val}}^b=\calL_{\calD_{val}}(\Theta^t,W^t)$\;
}
  Get  {\small $[\overline{\nabla}_{\Theta}\calL_S(\Theta^t,W^t); \overline{\nabla}_W \calL_S(\Theta^t,W^t)]$}  by~\cref{eq_asym_new_grad}\; 
  {\small $\overline{\delta}_{\Theta}^t, \overline{\delta}_W^t \leftarrow \emph{OP}([\overline{\nabla}_{\Theta}\calL_S(\Theta^t,W^t); \overline{\nabla}_W \calL_S(\Theta^t,W^t)])$}\;
   $\Theta^{t+1} = \emph{OP}(\eta, \Theta^t, \overline{\delta}_{\Theta}^t)$\;
  $W^{t+1} = \emph{OP}(\eta, W^t, \overline{\delta}_W^t)$\;
  $t = t + 1$\;
}
\Return{$\Theta^b, W^b$}
\end{minipage}
}
\end{algorithm}




\begin{table*}[t!]
    \centering
   \resizebox{0.9\textwidth}{!}{%
    \begin{tabular}{c | c c c c c c | c@{\hspace{2pt}} c@{\hspace{2pt}} c}
    \toprule
     Model & Texas & Wisconsin &Actor & Chameleon & Squirrel & Cornell & Roman\_Empire & Amazon\_Ratings \\
\toprule
ChebNet (S)&36.76\sd{6.76}&33.17\sd{7.83}&24.94\sd{1.33}&25.65\sd{3.56}&22.58\sd{2.56}&40.52\sd{8.15} &45.32\sd{0.65}&38.87\sd{0.26}\\
ChebNet (AS) &50.81\sd{8.03}&33.21\sd{5.33}&25.13\sd{0.83}&38.19\sd{1.26}&27.24\sd{1.95}&40.92\sd{7.92} &48.11\sd{0.41}&39.38\sd{0.19}\\
$\Delta \uparrow$ & \textcolor{blue}{+14.05}& \textcolor{blue}{+0.04}& \textcolor{blue}{+0.19}& \textcolor{blue}{+12.54}& \textcolor{blue}{+4.66}& \textcolor{blue}{+0.40}& \textcolor{blue}{+2.79}& \textcolor{blue}{+0.51}\\
\hline
ChebNetII (S)  &48.44\sd{10.87}&41.33\sd{9.25}&29.29\sd{0.8}&33.48\sd{4.59}&30.8\sd{2.65}&37.69\sd{11.1} &55.06\sd{0.32}&38.33\sd{0.42}\\
ChebNetII (AS)  &50.75\sd{10.0}&47.33\sd{10.17}&29.49\sd{0.9}&35.58\sd{3.7}&35.94\sd{2.66}&51.85\sd{6.01}&55.2\sd{0.3}&39.07\sd{0.28}\\
$\Delta \uparrow$& \textcolor{blue}{+2.31}& \textcolor{blue}{+6.00}& \textcolor{blue}{+0.20}& \textcolor{blue}{+2.10}& \textcolor{blue}{+5.14}& \textcolor{blue}{+14.16}& \textcolor{blue}{+0.14}& \textcolor{blue}{+0.74}\\
\hline

JacobiConv (S)  &50.17\sd{7.87}&46.08\sd{9.08}&32.66\sd{0.71}&26.57\sd{3.6}&23.15\sd{4.47}&49.71\sd{7.75}&52.92\sd{0.7}&40.18\sd{0.53}\\
JacobiConv (AS)  &52.66\sd{8.9}&49.5\sd{9.75}&32.78\sd{0.71}&29.95\sd{3.37}&24.71\sd{3.83}&52.83\sd{8.73}&54.16\sd{0.38}&47.17\sd{1.37}\\
$\Delta \uparrow$ & \textcolor{blue}{+2.49}& \textcolor{blue}{+3.42}& \textcolor{blue}{+0.12}& \textcolor{blue}{+3.38}& \textcolor{blue}{+1.56}& \textcolor{blue}{+3.12}& \textcolor{blue}{+1.24}& \textcolor{blue}{+6.99}\\
\hline
GPRGNN (S)  &48.55\sd{7.0}&40.79\sd{3.62}&30.2\sd{0.95}&30.44\sd{4.29}&24.33\sd{2.68}&46.53\sd{7.23}&55.48\sd{1.3}&39.86\sd{0.3}\\
GPRGNN (AS)  &49.54\sd{5.5}&43.46\sd{6.29}&30.75\sd{0.76}&31.63\sd{4.41}&24.82\sd{4.63}&47.63\sd{6.71}&55.99\sd{1.14}&39.89\sd{0.16}\\
$\Delta \uparrow$ & \textcolor{blue}{+0.99}& \textcolor{blue}{+2.67}& \textcolor{blue}{+0.55}& \textcolor{blue}{+1.19}& \textcolor{blue}{+0.49}& \textcolor{blue}{+1.10}& \textcolor{blue}{+0.51}& \textcolor{blue}{+0.03}\\
\hline
BernNet (S)  &52.14\sd{8.09}&49.33\sd{8.21}&30.57\sd{0.78}&29.45\sd{3.22}&25.94\sd{3.35}&48.15\sd{6.59}&55.3\sd{0.41}&39.33\sd{0.25}\\
BernNet (AS)  &53.87\sd{9.66}&52.46\sd{6.29}&30.8\sd{0.69}&30.46\sd{5.31}&27.5\sd{3.62}&49.88\sd{8.73}&55.5\sd{0.2}&39.78\sd{0.3}\\
$\Delta \uparrow$ & \textcolor{blue}{+1.73}& \textcolor{blue}{+3.13}& \textcolor{blue}{+0.23}& \textcolor{blue}{+1.01}& \textcolor{blue}{+1.56}& \textcolor{blue}{+1.73}&\textcolor{blue}{+0.20}& \textcolor{blue}{+0.45}\\
    \bottomrule
    \end{tabular}
   }
    \caption{Test accuracy with/without asymmetric learning on small heterophilic datasets (Texas, Wisconsin, Actor, Chameleon, Squirrel, Cornell) and large heterophilic datasets (Roman-Empire, Amazon-Ratings). 
    High accuracy indicates good performance. }
    \label{table-small-heter}
\end{table*}

\begin{table*}[ht]
    \centering
   \resizebox{0.9\textwidth}{!}{%
    \begin{tabular}{c@{\hspace{15pt}} | @{\hspace{20pt}}c@{\hspace{22pt}} c@{\hspace{22pt}} c@{\hspace{22pt}} c@{\hspace{22pt}} c@{\hspace{22pt}} c c }
    \toprule
   Model &Citeseer & Pubmed & Cora & Computers & Photo & Coauthor-CS & Coauthor-Physics\\
   \toprule

ChebNet (S)  &65.49\sd{1.39}&74.88\sd{1.47}&80.26\sd{0.86} & 78.76\sd{1.98}&90.26\sd{2.19}&90.74\sd{0.6}&94.69\sd{0.31}\\
ChebNet (AS)  &66.63\sd{1.1}&76.53\sd{1.52}&81.02\sd{0.92}&79.08\sd{2.6}&92.38\sd{0.41}&90.76\sd{0.73}&94.92\sd{0.1}\\
$\Delta \uparrow$ & \textcolor{blue}{+1.14}& \textcolor{blue}{+1.65}& \textcolor{blue}{+0.76} & \textcolor{blue}{+0.32}& \textcolor{blue}{+2.12}& \textcolor{blue}{+0.02}& \textcolor{blue}{+0.23}\\
\hline
ChebNetII (S)  &70.07\sd{1.13}&78.87\sd{1.28}&82.55\sd{0.78}&83.49\sd{0.9}&92.51\sd{0.43}&92.18\sd{0.23}&94.96\sd{0.13}\\
ChebNetII (AS)  &70.75\sd{0.78}&79.5\sd{0.94}&82.52\sd{0.95} &84.38\sd{1.3}&92.69\sd{0.43}&93.41\sd{0.23}&95.13\sd{0.09}\\
$\Delta \uparrow$ & \textcolor{blue}{+0.68}& \textcolor{blue}{+0.63} & -0.03 & \textcolor{blue}{+0.89}& \textcolor{blue}{+0.18}& \textcolor{blue}{+1.23}& \textcolor{blue}{+0.17}\\

\hline
JacobiConv (S)  &67.86\sd{1.08}&76.89\sd{1.67}&77.44\sd{1.23}&84.77\sd{0.7}&92.18\sd{0.42}&93.06\sd{0.3}&95.07\sd{0.16}\\
JacobiConv (AS)  &68.17\sd{0.9}&77.77\sd{1.37}&77.68\sd{1.24}&84.91\sd{0.46}&92.54\sd{0.27}&93.22\sd{0.31}&95.18\sd{0.16}\\
$\Delta \uparrow$ & \textcolor{blue}{+0.31}& \textcolor{blue}{+0.88}& \textcolor{blue}{+0.24} & \textcolor{blue}{+0.14}& \textcolor{blue}{+0.36}& \textcolor{blue}{+0.16}& \textcolor{blue}{+0.11}\\
\hline
GPRGNN (S)  &70.23\sd{1.29}&79.86\sd{1.81}&83.5\sd{0.94}&83.5\sd{0.99}&92.46\sd{0.29}&92.61\sd{0.19}&95.16\sd{0.08}\\ 
GPRGNN (AS)  &70.23\sd{1.03}&80.66\sd{1.64}&83.26\sd{0.69}&84.18\sd{0.71}&92.68\sd{0.43}&93.06\sd{0.49}&95.24\sd{0.13}\\
$\Delta \uparrow$ & \textcolor{blue}{+0.00}& \textcolor{blue}{+0.80} & -0.24& \textcolor{blue}{+0.68}& \textcolor{blue}{+0.22}& \textcolor{blue}{+0.45}& \textcolor{blue}{+0.08}\\
\hline
BernNet (S)  &68.64\sd{1.54}&79.58\sd{0.87}&82.37\sd{0.96}&85.1\sd{0.97}&92.65\sd{0.22}&92.3\sd{0.27}&95.29\sd{0.15}\\
BernNet (AS)  &69.46\sd{1.16}&79.62\sd{1.24}&82.47\sd{0.56}&
85.57\sd{0.96}&92.82\sd{0.36}&93.13\sd{0.37}&95.19\sd{0.18}\\
$\Delta \uparrow$ & \textcolor{blue}{+0.82}& \textcolor{blue}{+0.04}& \textcolor{blue}{+0.10} & \textcolor{blue}{+0.47}& \textcolor{blue}{+0.17}& \textcolor{blue}{+0.83}& -0.10\\
\bottomrule
    \end{tabular}
    }
    \caption{Test accuracy with/without asymmetric learning on homophilic graphs (Citeseer, Pubmed, Cora, Coauther-CS, Coauther-Physics, Amazon-Computer and Amazon-Photo).  
    High accuracy indicates good performance.}\vspace{-0.3cm}
    \label{table-homo}
\end{table*}

\paragraph{Smaller Block Condition Number.}

Preconditioning the gradient \(\nabla_{\Psi}\calL_S(\Psi^t) = [\nabla_{\Theta} \calL_S(\Theta^t,W^t), \nabla_W \calL_S(\Theta^t,W^t)]\) with  asymmetric preconditioner \(R^t\) is equivalent to preconditioning the Hessian matrix \(\bbH^t\) with \(R^t\) in the \(t\)-th iteration:
\begin{equation*}
    R^t \bbH^t  \Leftrightarrow R^t [\nabla_{\Theta} \calL_S(\Theta^t,W^t), \nabla_W \calL_S(\Theta^t,W^t)]
\end{equation*}
where \(\bbH_{ij}^t = \frac{\nabla_{\Psi_i} \calL_S(\Psi^t)}{\nabla_{\Psi_j} \calL_S(\Psi^t)}\), \((R^t \bbH^t)_{ij} = \frac{r_i^t \nabla_{\Psi_i} \calL_S(\Psi^t)}{\nabla_{\Psi_j} \calL_S(\Psi^t)}\), and \(r_i^t\) is the \(i\)-th diagonal element in \(R^t\). To simplify the analysis, we set 
\(\beta_{\pi_{\Theta}} = \beta_{\pi_W} = 0\) and consider the gradient descent optimizer~\citep{ema-book} for training.
We show that preconditioning with the asymmetric preconditioner results in a smaller block condition number of the Hessian when the following assumptions are met. The proofs are provided in detail in the Appendix.

\begin{restatable}[Proportional GPNR and Maximum Eigenvalue]{assumption}{propositionasymspeedhessian}\label{proposition-asym-speed-hessian}
Let \(\Psi\) and \(\Psi'\) be two points near a critical point \(\Psi^*\) such that \( \|\Psi - \Psi^* \|_2=\epsilon_1\) and \(\|\Psi' - \Psi^* \|_2=\epsilon_2\), where \(\epsilon_1 \rightarrow 0\) and \(\epsilon_2 \rightarrow 0\). Denote \(\bbH_{\Psi}\) and \(\bbH_{\Psi'}\) as the Hessian matrices of the empirical loss at points \(\Psi\) and \(\Psi'\), respectively. If \(\lambda_{\max}(\bbH_{\Psi}) \geq \lambda_{\max}(\bbH_{\Psi'})\), then \(\rho_{\Psi} \geq \rho_{\Psi'}\).
\end{restatable}

\begin{remark}
The relationship between the GPNR and the largest eigenvalue of Hessian is that \(\rho_{\Psi} \leq \lambda_{\max}(\bbH_{\Psi})\) when \(\Psi\) is near a critical point \(\Psi^*\) as stated in~\cref{proposition-asym-gpnr}.
Consequently, it is reasonable to assume that if the largest eigenvalues of the Hessian matrices at points \(\Psi\) and \(\Psi'\) satisfy \(\lambda_{\max}(\bbH_{\Psi}) \geq \lambda_{\max}(\bbH_{\Psi'})\), then the GPNRs at these points also satisfy \(\rho_{\Psi} \geq \rho_{\Psi'}\).
\end{remark}

\begin{restatable}[Mild Scaling]{assumption}{assumptionasymsscale}\label{assumption-asym-scale}
For the asymmetric preconditioner $R^t$ and the Hessian matrix $\bbH^t$, \(s_{\Theta}^t\) and \(s_W^t\) in \(R^t\) satisfy the inequality \(\frac{s_{\Theta}^t}{s_W^t} \geq \frac{\lambda_{max}(\bbH_{W,W}^t)}{\lambda_{max}(\bbH_{\Theta,\Theta}^t)}\).  
\end{restatable}

\begin{remark}
This assumption is also mild. The preconditioner is assumed to ensure that Proposition~\ref{proposition-asym-scale} holds. Specifically, it implies that the curvature of the loss function does not undergo significant changes after preconditioning.
\end{remark}

\begin{restatable}[Spectral Equivalent Preconditioner.]{proposition}{propositionasymsscale}\label{proposition-asym-scale}
Let $R^t$ be the asymmetric preconditioner and $\bbH^t$ the Hessian matrix at the $t$-th iteration.
If $\lambda_{max}(\bbH_{\Theta,\Theta}^t) \geq \lambda_{max}(\bbH_{W,W}^t)$ and $\frac{s_{\Theta}^t}{s_W^t}\geq \frac{ 
        \lambda_{max}(\bbH_{W,W}^t) }{ \lambda_{max}(\bbH_{\Theta,\Theta}^t) }$, we have $\lambda_{max}(\bbH'^t_{\Theta,\Theta}) \geq \lambda_{max}(\bbH'^t_{W,W})$ after preconditioning $\bbH'^t = R^t \bbH^t $.
\end{restatable}
\begin{remark}
    This proposition implies that the curvature of the loss function remains largely unchanged after preconditioning when the preconditioner $R^t$ satisfies certain conditions. The sensitivity of the parameters to the loss function is maintained after preconditioning. If the loss function is more sensitive to $\Theta$ than $W$, then $\Theta$ remains more sensitive than $W$ after preconditioning, and vice versa.
\end{remark}

\begin{restatable}[Smaller Block Condition Number.]{theorem}{theoremasymprecondition}\label{theorem-precondition} 
Given an asymmetric preconditioner $R^t$, the Hessian matrix $\bbH^t$ at the $t$-th iteration, and the gradient descent as the optimizer, under~\cref{assumption-asym-small-diff-norm},~\cref{proposition-asym-speed-hessian} and~\cref{assumption-asym-scale}, we have $\kappa'(\bbH'^t) \leq \kappa'(\bbH^t)$ after preconditioning $\bbH'^t = R^t \bbH^t$.

\end{restatable}

\begin{remark}
This theorem indicates that the block condition number decreases in each iteration during training when applying the asymmetric preconditioner. 
When the block condition number of the Hessian matrix approaches one, changes in 
$\Theta$ and $W$ lead to approximately the same changes in the empirical loss.
As the block condition number evaluates how poorly conditioned an optimization problem is, a small block condition number indicts an easy optimization problem.
\end{remark}

\section{Experiments}
\paragraph{Experimental Setup.}To empirically validate the effectiveness of our approach, we perform experiments on eighteen benchmark datasets for node classification tasks, including: six small heterophilic graphs (Texas, Wisconsin, Actor, Chameleon, Squirrel, Cornell), five large heterophilic graphs (Roman\_Empire, Amazon\_Ratings, Minesweeper, Tolokers, Questions) and seven homophilic graphs (Citeseer, Pubmed, Cora, Computers, Photo, Coauthor-CS, Coauthor-Physics). For each dataset, a sparse splitting strategy is employed where nodes are randomly partitioned into train/validation/test sets with proportions of $2.5\%/2.5\%/95\%$, respectively. Notably, for Citeseer, Pubmed, and Cora datasets, the setting from \cite{gprgnn,chebii} is adopted: 20 nodes per class are allocated for training, 500 nodes for validation, and 1,000 nodes for testing. 

We evaluate five prominent spectral GNN models as our baselines: ChebNet~\citep{chebynet}, GPRGNN~\citep{gprgnn}, BernNet~\citep{bernnet}, JacobiConv~\citep{jacobconv}, and ChebNetII~\citep{chebii}, and compare their performances in the setups with and without asymmetric learning across all datasets. For each GNN model, we use GNN (S) to denote the model trained using Adam optimizer~\citep{adam} and GNN (AS) to denote the model trained using Adam optimizer with asymmetric learning, with $\Delta \uparrow$ indicating the performance improvement.

Statistics of datasets, details about the baselines, and the setting of hyper-parameters are included in the Appendix.

\paragraph{Results and Discussion.} \cref{table-small-heter,table-homo} present the results for heterophilic graphs and homophilic graphs, respectively. Additional results for large heterophilic graphs can be further found in \cref{table-uniform-appendix-large-heter} in the Appendix.

\noindent\textbf{Q1: How does asymmetric learning perform with graphs of varying edge homophily ratios?}
We notice that asymmetric learning provides more performance gains on heterophilic graphs compared to homophilic ones.
As shown in \cref{table-small-heter}, on six small heterophilic datasets, there is an average accuracy improvement of \(3.08\%\), while an accuracy increase of approximately \(1.36\%\) is achieved on Romain\_Empire and Amazon\_Ratings.
In \cref{table-homo}, the average accuracy improvement is \(0.46\%\) on homophilic graphs. 
These results align with our theoretical analysis, showing that asymmetric learning facilitates the optimization of poorly conditioned problems by reducing the block condition number of the Hessian matrix. Empirically, we observe that the block condition numbers of Hessian matrices are larger on heterophilic graphs compared to homophilic graphs. Consequently, asymmetric learning yields a more significant improvement in classification accuracy on heterophilic graphs than on homophilic graphs.



\noindent\textbf{Q2: How does asymmetric learning perform with different spectral GNNs?}
Spectral GNNs can be categorized based on their polynomial basis into orthogonal spectral GNNs (ChebNet, ChebNetII, and JacobiConv) and non-orthogonal spectral GNNs (GPRGNN and BernNet).
We observe that asymmetric learning generally exhibits superior performance with orthogonal spectral GNNs compared to non-orthogonal spectral GNNs. For example, there is a \(5.15\%\) accuracy improvement with orthogonal spectral GNNs using asymmetric learning on the six small heterophilic datasets (\cref{table-small-heter}), whereas non-orthogonal spectral GNNs show a \(1.69\%\) accuracy improvement. 
A plausible explanation is that the interference between different orders of orthogonal polynomial bases is significantly lower than that between non-orthogonal polynomial bases. Suppose that \(\Theta=\{\theta_k\}_{k=0}^K\), where each \(\theta_k\) is associated with a polynomial basis \(T_k\) of order \(k\). Due to the interference in non-orthogonal polynomial bases, a sophisticated preconditioner is required, rather than simply scaling all \(\theta_k\) uniformly by \(s_{\Theta}\), to achieve better performance.


\begin{figure}[htbp!]
    \centering   \includegraphics[width=0.8\columnwidth ]{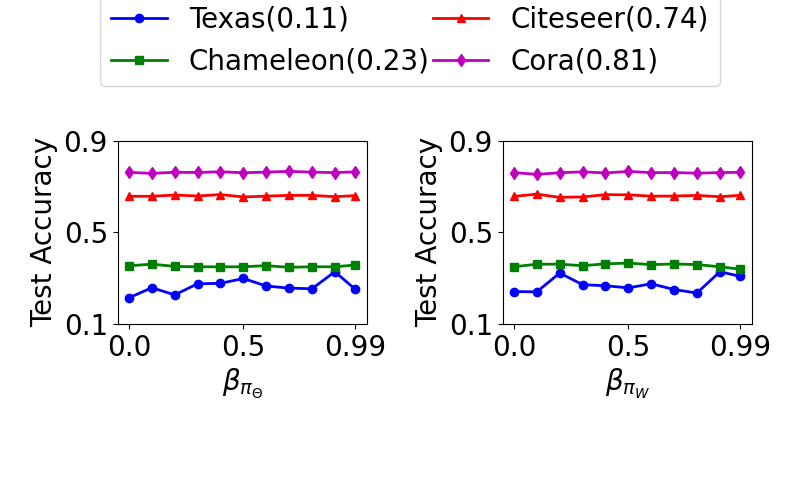}\vspace{-0.7cm}  
    \caption{The test accuracy of ChebNet when applying asymmetric learning with varying values of $\beta_{\pi_{\Theta}}$ and $ \beta_{\pi_W}$.}
    \label{fig_beta}\vspace{-0.3cm}
\end{figure}

\vspace{0.1cm}
\noindent\textbf{Q3: How does asymmetric learning perform with graphs of different sizes?} Asymmetric learning shows greater performance improvement on datasets with smaller graph sizes. In \cref{table-small-heter}, there is an average accuracy improvement of \(3.08\%\) on smaller graphs on the left. For larger graphs on the right, improvements are smaller, with a \(1.36\%\) increase in accuracy. Specifically, 
spectral GNNs achieve test accuracy improvements exceeding \(10\%\) on smaller datasets such as Texas, Wisconsin, Cornell, and Chameleon, which contain fewer than 1,000 nodes. 
As smaller training data sizes can often lead to instability in the gradient of empirical loss during training, these findings suggest that Asymmetric learning mitigates this issue by adjusting parameter updates without risking overly aggressive updates. 

\noindent\textbf{Q4: How do the exponential decay parameters $\beta_{\pi_{\Theta}}$ and $\beta_{\pi_W}$ affect the performance of asymmetric learning?}
We explore the impact of the hyperparameters $\beta_{\pi_{\Theta}}$ and $\beta_{\pi_W}$ on test accuracy when using ChebNet for node classification tasks. The parameters $\beta_{\pi_{\Theta}}$ and $\beta_{\pi_W}$ increase from 0 to 0.9 in steps of 0.1, with an additional value at 0.99. Figure~\ref{fig_beta} illustrates that fluctuations in performance are observed only in the Texas dataset, while marginal differences are noted in other datasets. This suggests that the performance is not sensitive to $\beta_{\pi_{\Theta}}$ and $\beta_{\pi_W}$ when training samples are sufficient. Maintaining a training data splitting ratio of \(2.5\%\) can ensure an adequate amount of training data for large graphs.

\section{Related Works}

\paragraph{Spectral GNNs and Optimization}
Spectral GNNs transform graph signals from the vertex domain to the spectral domain for filtering and then applying an inverse transformation to return the filtered signals to the vertex domain \citep{chebynet}. Based on the design of their graph filters, spectral GNNs are categorized into polynomial and rational types. Polynomial spectral GNNs employ various polynomial basis to approximate graph filters. For example, ChebNet and ChebNetII utilize Chebyshev polynomial basis~\citep{chebynet,chebii}, while JacobiConv uses Jacobi polynomial basis \citep{jacobconv}. These models leverage orthogonal polynomial basis, which allow efficient approximation as approximation error decreases exponentially with increasing polynomial order. Non-orthogonal polynomial spectral GNNs are also noteworthy, such as BernNet~\citep{bernnet} which uses Bernstein polynomial basis, and GPRGNN~\citep{gprgnn} which employs monomial polynomial basis. 
Rational spectral GNNs 
offer greater flexibility at the cost of increased complexity. For instance, CayleyNet applies parametric rational functions to identify relevant frequency bands~\citep{cayleynet}; ARMA incorporates moving average terms to enhance noise robustness and capture long-term dynamics on graphs~\citep{arma}. LanczosNet facilitates multi-scale analysis of graph signals~\citep{lancz}. 
Polynomial spectral GNNs are computationally efficient and enable localized vertex analysis~\citep{wavelets}. This paper focuses on polynomial spectral GNNs.


A crucial area in spectral GNNs is the optimization. Studies on optimization techniques for GNNs are very limited. \citet{Gnn-OptimizationOG} have investigated the effects of network depth and skip connections on the convergence behavior of linear GNNs. \citet{natural-gradient} have demonstrated the use of natural gradient methods during training to produce more effective solutions. A common practice in optimizing spectral GNNs is to apply different learning rates to graph convolution parameters and feature transformation parameters~\citep{bernnet,chebii}. Nevertheless, the rationale behind this practice remains inadequately explored.


\vspace{0.1cm}

\noindent\textbf{Learning Rate Selection and Preconditioning}
The optimization of neural networks heavily depends on the choices of learning rates for parameter updates. Studies emphasize that the learning rate should not exceed the reciprocal of the largest eigenvalue of the Hessian matrix to prevent overshooting and convergence to inferior solutions in gradient descent algorithms~\citep{rate_eigenvalue-2,overshoot-1}. 

The optimization of spectral GNNs is further complicated by the distinct characteristics of graph convolution parameters $\Theta$ and feature transformation parameters $W$, leading to a poorly conditioned problem.
A popular approach to tackle poor-conditioned optimization problems is \emph{preconditioning}~\citep{Saad2003iterative}. First-order methods such as AdaGrad~\citep{AdaGrad}, RMSProp~\citep{RMSProp}, and Adam~\citep{adam} all incorporate a preconditioning component. These methods adapt learning rates based on historical gradients, assuming that all parameters and their gradients are of the same magnitude.  
Low-rank preconditioners are widely used in the large-scale matrix factorization and deep neural architectures~\citep{scaledGD,shampoo,block-low-rank}. However, there are no works specifically tailored to address the asymmetric parameters in spectral GNNs.


\section{Conclusion and Future Work}
We have demonstrated that the parameters 
$\Theta$ and $W$ in spectral GNNs play distinct roles, resulting in a poorly conditioned optimization problem. To address this, we introduced the concept of the block condition number to estimate the optimization difficulty and proposed asymmetric learning, a novel preconditioning technique designed to facilitate the optimization process. Our experiments validate the efficacy of asymmetric learning, particularly in the optimization on heterophilic graphs. These findings underscore the potential of asymmetric learning to enhance the performance and training efficiency of spectral GNNs across diverse graph datasets. Future work will explore further applications of this approach to other types of GNNs and graph-based learning tasks.


\clearpage
\subsection*{Acknowledgements}
We are thankful to Yun Kuen Cheung for helpful discussions. 
This research was supported partially by the Australian Government through the Australian Research Council's Discovery Projects funding scheme (project DP210102273).



\clearpage
\appendix
\appendix



\section{Proofs}\label{appendix-asym-proof}
\begin{lemma}[Weyl Inequality~\citep{weyl-ineq}]\label{inequality_weyl}
Let $A$ and $B$ be two Hermitian matrices and $\lambda_{max}(\cdot)$ denote the maximum eigenvalue of a matrix. The following holds:
\begin{equation}
    \lambda_{max}(A+B)\leq \lambda_{max}(A) + \lambda_{max}(B). 
\end{equation}

\end{lemma}
We refer the readers to \citep{weyl-ineq} for the proof. 

\propositionasymgpnr*
\begin{proof}
    Let \(\Psi\) be a parameter of the function \(\calL_S\) near a critical point \(\Psi^*\) such that \(\epsilon= \Psi - \Psi^*, \|\epsilon\|_2 \rightarrow 0\) and \(\nabla_{\Psi} \calL_S(\Psi^*)=0\). By the Taylor expansion, we have
    \begin{equation}\label{eq_asym_psi_taylor}
        \calL_S(\Psi) = \calL_S(\Psi^*) + \frac{1}{2}\epsilon^T \bbH_{\Psi^*} \epsilon + O(\|\epsilon\|_2^3),
    \end{equation}
    where \(\bbH_{\Psi^*}\) denotes the Hessian matrix at \(\Psi^*\) and $O(\|\epsilon\|_2^3)$ the remainder term captures the error introduced by truncating the Taylor series after the second-order term.

    By differentiating both sides of~\cref{eq_asym_psi_taylor} with respect to \(\Psi\), we obtain
    \begin{equation}\label{eq_asym_appendix_1}
        \nabla_{\Psi} \calL_S(\Psi) = \bbH_{\Psi^*} \epsilon  + O( 3\|\epsilon\|_2 \epsilon  ).
    \end{equation}
    
    By differentiating again with respect to \(\Psi\), we obtain
    \begin{equation}\label{eq_asym_appendix_hessian}
        \bbH_{\Psi} =  \bbH_{\Psi^*} + O\left(  3 * \left( \frac{\epsilon \epsilon^T}{\|\epsilon\|_2}  + \|\epsilon\|_2  I \right)\right).
    \end{equation}

    By the definition of eigenvalue, for any vector \(\Psi\), we have
    \begin{equation}\label{eq_asym_appendix_2}
        \|\bbH_{\Psi} \epsilon \|_2 \leq \lambda_{max}(\bbH_{\Psi}) \| \epsilon\|_2.
    \end{equation}
    
    This leads to the following:
    
    \begin{align*}
    \begin{split}
 &\frac{\|\nabla_{\Psi} \calL_S(\Psi)\|_2}{\|\Psi\|_2} = \frac{\| \bbH_{\Psi^*} \epsilon + O(3 \|\epsilon\|_2 \epsilon) \|_2}{\|\Psi\|_2} \\
    &= \frac{\|\left( \bbH_{\Psi} - O\left( 3 \left( \frac{\epsilon \epsilon^T}{\|\epsilon\|_2} + \|\epsilon\|_2 I \right) \right)\right) \epsilon + O(3 \|\epsilon\|_2 \epsilon) \|_2}{\|\Psi\|_2} \\
        &\leq \frac{\|\bbH_{\Psi} \epsilon \|_2}{\|\Psi \|_2} + \frac{\| O\left( 3 \left( \frac{\epsilon \epsilon^T}{\|\epsilon\|_2} + \|\epsilon\|_2 I \right) \right) \epsilon \|_2}{\|\Psi \|_2} \\
            &\quad + \frac{\| O(3 \|\epsilon\|_2 \epsilon) \|_2}{\|\Psi \|_2}\\
                &\leq \frac{\lambda_{\max}(\bbH_{\Psi}) \|\epsilon\|_2}{\|\Psi \|_2} + \frac{O\left( \lambda_{\max}\left( 3 \left( \frac{\epsilon \epsilon^T}{\|\epsilon\|_2} + \|\epsilon\|_2 I \right) \right) \|\epsilon \|_2 \right)}{\|\Psi \|_2} \\
    &\quad + \frac{O(3 \|\epsilon \|_2^2)}{\|\Psi \|_2}, \quad~\text{\cref{eq_asym_appendix_2}} 
    \end{split}
    \end{align*}  \begin{align}\label{eq_asym_appendix_rho_eigen}
    \begin{split}
    &\leq \frac{\lambda_{\max}(\bbH_{\Psi}) \|\Psi\|_2}{\|\Psi \|_2} + \frac{O\left( \lambda_{\max}\left( 3 \left( \frac{\epsilon \epsilon^T}{\|\epsilon\|_2} + \|\epsilon\|_2 I \right) \right) \|\epsilon \|_2 \right)}{\|\Psi \|_2} \\
    &\quad + \frac{O(3 \|\epsilon \|_2^2)}{\|\Psi \|_2}, \quad~\text{\cref{assumption-asym-small-diff-norm}}\\
    &= \lambda_{\max}(\bbH_{\Psi}) + \frac{O\left( \lambda_{\max}\left( 3 \left( \frac{\epsilon \epsilon^T}{\|\epsilon\|_2} + \|\epsilon\|_2 I \right) \right) \|\epsilon \|_2 \right)}{\|\Psi \|_2} \\
    &\quad + \frac{O(3 \|\epsilon \|_2^2)}{\|\Psi \|_2}
    \end{split}
    \end{align}


The rank of matrix $\epsilon \epsilon^T$ is $1$ and it has one nonzero eigenvalue $\|\epsilon\|_2^2$. All eigenvalues of the diagonal matrix $\|\epsilon\|_2 I$ are $ \|\epsilon\|_2$, i.e.,
\begin{equation}\label{eq_asym_appendix_max}
    \lambda_{max}(\epsilon \epsilon^T)= \|\epsilon\|_2^2;\quad \lambda_{max}(\|\epsilon\|_2 I)=\|\epsilon\|_2.
\end{equation}

This leads to the following

\begin{align}\label{eq_asym_appendix_eig_term1}
    \begin{split}
        &\lambda_{\max}\left( 3 \left( \frac{\epsilon \epsilon^T}{\|\epsilon\|_2} + \|\epsilon\|_2 I \right) \right)\\
        &\leq 3\left(\lambda_{\max}\left( \frac{\epsilon \epsilon^T}{\|\epsilon\|_2}\right) + \lambda_{\max}\left( \|\epsilon\|_2 I\right)   \right),\quad ~\cref{inequality_weyl}\\
        &=3\left( \frac{1}{\|\epsilon\|_2} \lambda_{\max}\left(\epsilon \epsilon^T \right) + \lambda_{\max}\left( \|\epsilon\|_2 I\right)\right)\\
        &=3\left(\frac{1}{\|\epsilon\|_2} * \|\epsilon\|_2^2 +  \|\epsilon\|_2\right),\quad~\cref{eq_asym_appendix_max}\\
        &= 6\|\epsilon\|_2
    \end{split}
\end{align}

By substituting~\cref{eq_asym_appendix_eig_term1} into~\cref{eq_asym_appendix_rho_eigen}, we have
\begin{align*}
    \begin{split}
        \frac{\|\nabla_{\Psi} \calL_S(\Psi)\|_2}{\|\Psi\|_2}&\leq  \lambda_{\max}(\bbH_{\Psi}) + 
        \frac{O\left( 6\|\epsilon\|_2^2 \right)}{\| \Psi\|_2} + 
        \frac{O\left( 3\|\epsilon\|_2^2 \right)}{\| \Psi\|_2}\\
        &=\lambda_{\max}(\bbH_{\Psi}) + 
        \frac{O\left( 9\|\epsilon\|_2^2 \right)}{\| \Psi\|_2} 
    \end{split}
\end{align*}

When $\|\epsilon \|_2\rightarrow 0$,  we have
   \[
   \rho_{\Psi}= \frac{\|\nabla_{\Psi} \calL_S(\Psi)\|_2}{\|\Psi\|_2} \leq \lambda_{max}(\bbH_{\Psi}).
   \]

\end{proof}

\propositionasymsscale*
\begin{proof}
    After preconditioning \(\bbH^t\) with the preconditioner \(R^t\), we have
    \begin{align*}
        \begin{split}
            \lambda_{max}(\bbH'^t_{\Theta,\Theta}) &= s_{\Theta} \lambda_{max}(\bbH^t_{\Theta,\Theta});\\
             \lambda_{max}(\bbH'^t_{W,W}) &= s_W \lambda_{max}(\bbH^t_{W,W}).
        \end{split}
    \end{align*}

    This yields
    \begin{equation*}
         \frac{\lambda_{max}(\bbH'^t_{\Theta,\Theta})}{\lambda_{max}(\bbH'^t_{W,W})} = \frac{s_{\Theta}}{s_W} \frac{\lambda_{max}(\bbH^t_{\Theta,\Theta})}{\lambda_{max}(\bbH^t_{W,W})}.
    \end{equation*}
    Then, when \(\frac{s_{\Theta}}{s_W} \geq \frac{\lambda_{max}(\bbH^t_{W,W})}{\lambda_{max}(\bbH^t_{\Theta,\Theta})}\), we obtain
    \begin{equation*}
        \frac{\lambda_{max}(\bbH'^t_{\Theta,\Theta})}{\lambda_{max}(\bbH'^t_{W,W})} \geq 1.
    \end{equation*}
    Therefore, \(\lambda_{max}(\bbH'^t_{\Theta,\Theta}) \geq \lambda_{max}(\bbH'^t_{W,W})\) is derived.
\end{proof}

\theoremasymprecondition*
\begin{proof}
    (1) We first prove that, under~\cref{assumption-asym-small-diff-norm}, the following holds for the GPNRs of $\Theta$ and $W$ and the largest eigenvalues of block Hessian matrices $\bbH_{\Theta,\Theta}$ and $\bbH_{W,W}$: $\rho_{\Theta}\leq \lambda_{max}(\bbH_{\Theta,\Theta})$ and $\rho_{W,W}\leq \lambda_{max}(\bbH_{W,W})$.
    

    For any parameter \(\Psi = \{\Theta, W\}\) around a critical point \(\Psi^* = \{\Theta^*, W^*\}\) such that \( \epsilon_{\Theta} = \Theta - \Theta^*, \|\epsilon_{\Theta}\|_2\rightarrow 0\) and \(\epsilon_W = W - W^*\) and \(\|\epsilon_W\|_2\rightarrow 0\), the second-order Taylor expansion with the remainder term for $\Theta$ is
    \begin{align}\label{eq_asym_appendix_taylor_theta}
        \begin{split}
            \mathcal{L}_S(\Theta, W) &= \mathcal{L}_S(\Theta^*, W^*) + \nabla_{\Theta} \mathcal{L}_S(\Theta^*, W^*)^T\epsilon_{\Theta} \\
            &+
    \frac{1}{2}\epsilon_{\Theta}^T \bbH_{\Theta^*,\Theta^*} \epsilon_{\Theta} + O(\|\epsilon_{\Theta}\|_2^3).
        \end{split}
    \end{align}

    By differentiating both side of~\cref{eq_asym_appendix_taylor_theta} with respect to $\Theta$, we have
    \begin{equation}\label{eq_asym_appendix_taylor_theta_first}
        \nabla_{\Theta} \calL_S(\Theta,W) =   \bbH_{\Theta^*,\Theta^*}\epsilon_{\Theta} + O(3\|\epsilon_{\Theta}\|_2 \epsilon_{\Theta}).
    \end{equation}
    
    By differentiating both side of~\cref{eq_asym_appendix_taylor_theta} twice with respect to $\Theta$, we obtain 
    \begin{equation}\label{eq_asym_appendix_taylor_theta_second}
        \bbH_{\Theta,\Theta} =   \bbH_{\Theta^*,\Theta^*} + O\left(   3 * \left( \frac{\epsilon_{\Theta} \epsilon_{\Theta}^T}{\|\epsilon_{\Theta}\|_2}  + \|\epsilon_{\Theta}\|_2 * I\right) \right).
    \end{equation}

    Repeat steps in~\cref{eq_asym_appendix_rho_eigen}, when $\|\epsilon_{\Theta} \|_2\rightarrow 0$, we have
    \begin{equation}
        \rho_{\Theta}=\frac{\|\nabla_{\Theta} \calL_S(\Theta,W)\|_2 }{\|\Theta\|_2 }\leq  \lambda_{max}(\bbH_{\Theta,\Theta}).
    \end{equation}

    By repeating Taylor expansion at $\Psi^*=\{\Theta^*,W^*\}$ for $\epsilon_W=W -W^*$ and above steps, when $\|\epsilon_W\|_2\rightarrow 0$, we have
    \begin{equation}
        \rho_W=\frac{\|\nabla_W \calL_S(\Theta,W)\|_2 }{\|W\|_2 }\leq  \lambda_{max}(\bbH_{W,W}).
    \end{equation}

    (2) Next, we show that under~\cref{proposition-asym-speed-hessian} and~\cref{assumption-asym-scale}, the block condition number decreases after preconditioning.

    For the Hessian matrix \(\bbH^t\) and its preconditioned form \(\bbH'^t = R^t \bbH^t \), where \(R^t\) is the asymmetric preconditioner with the diagonal elements \(s_{\Theta}^t\) and \(s_W^t\) set according to \cref{eq_asym_grad_scale}. 
    By~\cref{assumption-asym-scale},
    if \(\lambda_{\max}(\bbH_{\Theta, \Theta}^t) \geq \lambda_{\max}(\bbH_{W, W}^t)\), 
    we have 
    \(\lambda_{\max}(\bbH'^t_{\Theta, \Theta}) \geq \lambda_{\max}(\bbH'^t_{W, W})\) after preconditioning.
    Thus, the block condition number of \(\bbH'^t\) is obtained as
    \begin{align}\label{eq_asym_appendix_tmp1}
        \begin{split}
            \kappa'(\bbH'^t) &= \frac{s_{\Theta}^t \lambda_{max}(\bbH_{\Theta, \Theta}^t)}{s_W^t \lambda_{max}(\bbH_{W, W}^t)} \\
            &= \frac{s_{\Theta}^t}{s_W^t} \kappa'(\bbH^t).
        \end{split}
    \end{align}

    According to~\cref{proposition-asym-speed-hessian}, when \(\lambda_{\max}(\bbH_{\Theta, \Theta}^t) \geq \lambda_{\max}(\bbH_{W, W}^t)\), we have \(\rho_{\Theta}^t \geq \rho_W^t\), i.e., \(\frac{\|\nabla_{\Theta} \calL_S(\Theta^t, W^t)\|_2}{\|\Theta^t\|_2} \geq \frac{\|\nabla_W \calL_S(\Theta^t, W^t)\|_2}{\|W^t\|_2}\). When \(s_{\Theta}^t\) and \(s_W\) are set according to \cref{eq_asym_grad_scale}, we have
    \begin{align}\label{eq_asym_appendix_tmp2}
        \begin{split}
            \frac{s_{\Theta}^t}{s_W^t} &= \frac{\frac{\|\Theta^t\|_2}{\|\nabla_{\Theta} \calL_S(\Theta^t, W^t)\|_2}}{\frac{\|W^t\|_2}{\|\nabla_W \calL_S(\Theta^t, W^t)\|_2}} \\
            &= \frac{\frac{\|\nabla_W \calL_S(\Theta^t, W^t)\|_2}{\|W^t\|_2}}{\frac{\|\nabla_{\Theta} \calL_S(\Theta^t, W^t)\|_2}{\|\Theta^t\|_2}} \\
            &\leq 1.
        \end{split}
    \end{align}

    By substituting \cref{eq_asym_appendix_tmp2} into \cref{eq_asym_appendix_tmp1}, we obtain
    \begin{equation*}
        \kappa'(\bbH'^t) \leq \kappa(\bbH^t).
    \end{equation*}
\end{proof}


\begin{table*}[htbp!]
    \centering
    \resizebox{0.65\textwidth}{!}{
   \begin{tabular}{*{7}{c}}
   \toprule
  Statistics & Texas & Wisconsin & Cornell & Actor & Chameleon & Squirrel \\ 
\toprule
        \# Nodes & 183 & 251 & 183 & 7,600 & 890 & 2,223  \\
        \# Edges & 295 & 466 & 280 & 26,752 & 27,168 & 131,436 \\
        \# Features & 1,703 & 1,703 & 1,703 & 932 & 2,325 & 2,089  \\
        \# Classes & 5 & 5 & 5 & 5 & 5 & 5  \\
        \# Edge Homophily & 0.11 & 0.21 & 0.3 & 0.22 & 0.24 & 0.22  \\
\bottomrule
\end{tabular}
}
\caption{Statistics of six small heterophilic datasets~\citep{GeomGCNGG,multiscale,critical-heter}. }\vspace{0.4cm}
\resizebox{0.7\textwidth}{!}{
\begin{tabular}{*{6}{c}}
\toprule
Statistics& Roman\_Empire & Amazon\_Ratings & Tolokers & Minesweeper & Questions \\
        \toprule
        \# Nodes & 22,662 & 24,492 & 11,758 & 10,000 & 48,921 \\
        \# Edges & 32,927 & 93,050 & 519,000 & 39,402 & 153,540 \\
        \# Features & 300 & 300 & 10 & 7 & 301 \\
        \# Classes & 18 & 5 & 2 & 2 & 2 \\
        \# Edge Homophily & 0.05 & 0.38 & 0.59 & 0.68 & 0.84 \\
\bottomrule
\end{tabular}

}
\caption{Statistics of five large heterophilic datasets~\cite{critical-heter}.}\vspace{0.4cm}
\resizebox{0.8\textwidth}{!}{
\begin{tabular}{*{8}{c}}
\toprule
Statistics&Citeseer & Pubmed & Cora & Computers & Photo & Coauthor-CS & Coauthor-Physics \\
        \toprule
        \# Nodes & 3,327 & 19,717 & 2,708 & 13,752 & 7,650 & 18,333 & 134,493 \\
        \# Edges & 4,676 & 44,327 & 5,278 & 491,722 & 238,162 & 163,788 & 495,924 \\
        \# Features & 3,703 & 500 & 1,433 & 767 & 745 & 6,805 & 8,415 \\
        \# Classes & 6 & 5 & 7 & 10 & 8 & 15 & 5 \\
        \# Edge Homophily & 0.74 & 0.8 & 0.81 & 0.78 & 0.83 & 0.81 & 0.93 \\
\bottomrule
\end{tabular}
}
\caption{Statistics of homophilic datasets, including three small datasets (Citeseer, Pubmed, Cora) and four large datasets (Computers, Photo, Coauthor-CS, Coauthor-Physics)~\citep{gcn,graphsaint,coauthor}. }
    \label{table_dataset}
\end{table*}

\begin{table*}[ht]
    \centering
    \resizebox{0.75\textwidth}{!}{%
    \begin{tabular}{c |c c c c c}
    \toprule
      GNNs&   
      Roman\_Empire & Amazon\_Ratings & Minesweeper &  Tolokers & Questions\\
      \toprule
      ChebNet (S)  &45.32\sd{0.65}&38.87\sd{0.26}&86.34\sd{0.19}&69.88\sd{0.26}&48.55\sd{1.04}\\
ChebNet (AS)  &48.11\sd{0.41}&39.38\sd{0.19}&86.71\sd{0.15}&70.05\sd{0.36}&51.84\sd{0.31}\\
$\Delta \uparrow$ & \textcolor{blue}{+2.79}& \textcolor{blue}{+0.51}& \textcolor{blue}{+0.37}& \textcolor{blue}{+0.17}& \textcolor{blue}{+3.29}\\
\hline
ChebNetII (S)   &55.06\sd{0.32}&38.33\sd{0.42}&74.49\sd{3.76}&69.37\sd{0.6}&63.99\sd{0.32}\\
ChebNetII (AS)   &55.2\sd{0.3}&39.07\sd{0.28}&78.16\sd{0.1}&69.65\sd{0.81}&63.68\sd{0.81}\\
$\Delta \uparrow$ & \textcolor{blue}{+0.14}& \textcolor{blue}{+0.74}& \textcolor{blue}{+3.67}& \textcolor{blue}{+0.28}& -0.31\\
\hline
JacobiConv (S)   &52.92\sd{0.7}&40.18\sd{0.53}&87.4\sd{0.13}&70.24\sd{0.18}&55.68\sd{0.54}\\
JacobiConv (AS)  &54.16\sd{0.38}&47.17\sd{1.37}&87.66\sd{0.11}&70.08\sd{0.28}&56.25\sd{0.65}\\
$\Delta \uparrow$ & \textcolor{blue}{+1.24}& \textcolor{blue}{+6.99}& \textcolor{blue}{+0.26}& -0.16& \textcolor{blue}{+0.57}\\
\hline
GPRGNN (S)    &55.48\sd{1.3}&39.86\sd{0.3}&86.68\sd{0.32}&67.05\sd{0.97}&53.76\sd{0.41}\\
GPRGNN (AS)    &55.99\sd{1.14}&39.89\sd{0.16}&87.11\sd{0.39}&68.22\sd{1.1}&56.18\sd{0.3}\\
$\Delta \uparrow$ & \textcolor{blue}{+0.51}& \textcolor{blue}{+0.03}& \textcolor{blue}{+0.43}& \textcolor{blue}{+1.17}& \textcolor{blue}{+2.42}\\
\hline
BernNet (S)  &55.3\sd{0.41}&39.33\sd{0.25}&76.64\sd{0.29}&69.31\sd{0.69}&65.41\sd{0.33}\\
BernNet (AS)  &55.5\sd{0.2}&39.78\sd{0.3}&77.02\sd{0.17}&70.39\sd{0.62}&64.88\sd{0.31}\\
$\Delta \uparrow$ & \textcolor{blue}{+0.20}& \textcolor{blue}{+0.45}& \textcolor{blue}{+0.38}& \textcolor{blue}{+1.08}& -0.53\\
    \bottomrule
    \end{tabular}
    }
    \caption{Performance of GNNs with/without asymmetric learning on large heterophilic datasets (Roman-Empire, Amazon-Ratings, Minesweeper, Tolokers, Questions) when data splitting is $2.5\%/2.5\%/95\%$. Test accuracy is used as the metric for Roman-Empire and Amazon-Ratings datasets and ROC AUC is reported on Minesweeper, Tolokers, Questions. High accuracy and ROC AUC indict good performance. }
    \label{table-uniform-appendix-large-heter}
\end{table*}

\section{Spectral GNNs}
We provide the detailed description for spectral GNNs used in our experiments in the following.

For a graph with the adjacency matrix $A$, the degree matrix $D$, and the identity matrix $I$, we use $\hat{L}=I-D^{-1/2}AD^{-1/2}$, $\tilde{L}=-D^{-1/2}AD^{-1/2}$, $\tilde{A}=D^{-1/2}AD^{-1/2}$, and $\tilde{A}'=(D+I)^{-1/2}(A+I)(D+I)^{-1/2}$ to denote the normalized Laplacian matrix, the shifted normalized Laplacian matrix, the normalized adjacency, matrix and the normalized adjacency matrix with self-loops, respectively.

\bigskip
\noindent\textbf{ChebNet} \citep{chebynet}: This model uses the Chebyshev basis to approximate a spectral filter:
\begin{equation*}
    \hat{Y}=\sum_{k=0}^K \theta_k T_k(\tilde{L}) f_W(X)
\end{equation*}
where $X$ is the raw feature matrix, $\Theta=[\theta_0,\theta_1,\ldots, \theta_K]$ is the graph convolution parameter, $W$ is the feature transformation parameter and $f_W(X)$ is usually a 2-layer MLP. $T_k(\tilde{L})$ is the $k$-th Chebyshev basis expanded on the shifted normalized graph Laplacian matrix $\tilde{L}$ and is recursively calculated: 
\begin{align*}
    T_0(\tilde{L})&= I\\
    T_1(\tilde{L})&=\tilde{L}\\
    T_k(\tilde{L})&=2\tilde{L}T_{k-1}(\tilde{L})-T_{k-2}(\tilde{L})
\end{align*}
\bigskip

\noindent\textbf{ChebNetII} \citep{chebii}: The model is formulated as
\begin{equation*}
    \hat{Y} = \frac{2}{K+2}{}\sum_{k=0}^K \sum_{j=0}^K \theta_j T_k(x_j)T_k(\tilde{L}) f_W(X),
\end{equation*}
where $X$ is the input feature matrix, $W$ is the feature transformation parameter, $f_W(X)$ is usually a 2-layer MLP, $T_k(\cdot)$ is the $k$-th Chebyshev basis expanded on $\cdot$, $x_j = \cos \left( \left(j+1/2 \right)\pi /\left(K+1 \right) \right)$ is the $j$-th Chebyshev node, which is the root of the Chebyshev polynomials of the first kind with degree $K+1$, and $\theta_j$ is a learnable parameter. Graph convolution parameter in ChebNet is reparameterized with Chebyshev nodes and learnable parameters $\theta_j$. 

\bigskip

\noindent\textbf{JacobiNet} \citep{jacobconv}: This model uses the Jacobi basis to approximate a filter as:
\begin{equation*}  
    \hat{Y} = \sum_{k=0}^K \theta_k P_k^{a,b}(\tilde{A}) f_W(X),
\end{equation*}
where $X$ is the input feature matrix, $\Theta=[\theta_0,\theta_1,\ldots, \theta_K]$ is the graph convolution parameter, $W$ is the feature transformation parameter and $f_W(X)$ is usually a 2-layer MLP. $P_k^{a,b}(\tilde{A})$ is the Jacobi basis on normalized graph adjacency matrix $\tilde{A}$ and is recursively calculated as
\begin{align*}
    P_k^{a,b}(\tilde{A})&=I\\
    P_k^{a,b}(\tilde{A})&=\frac{1-b}{2}I + \frac{a+b+2}{2}\tilde{A}\\
    P_k^{a,b}(\tilde{A})&= \gamma_k \tilde{A} P_{k-1}^{a,b}(\tilde{A}) + \gamma'_k P_{k-1}^{a,b}(\tilde{A}) + \gamma''_k P_{k-2}^{a,b}(\tilde{A})
\end{align*}
where $\gamma_k = \frac{(2k+a+b)(2k+a+b-1)}{2k(k+a+b)}, \gamma'_k = \frac{(2k+a+b-1)(a^2-b^2)}{2k(k+a+b)(2k+a+b-2)},\gamma''_k = \frac{(k+1-1)(k+b-1)(2k+a+b)}{k(k+a+b)(2k+a+b-2)}$. $a$ and $b$ are hyper-parameters. Usually, grid search is used to find the optimal $a$ and $b$ values.

\bigskip


\noindent\textbf{GPRGNN} \citep{gprgnn}: This model uses the monomial basis to approximate a filter:
\begin{equation*}
    \hat{Y} = \sum_{k=0}^K \theta_k \tilde{A}'^k f_W(X)
\end{equation*}
where $X$ is the input feature matrix, $\Theta=[\theta_0,\theta_1,\ldots, \theta_K]$ is the graph convolution parameter, $W$ is the feature transformation parameter and $f_W(X)$ is usually a 2-layer MLP. $\tilde{A}'$ is the normalized adjacency matrix with self-loops.
\bigskip

\noindent\textbf{BernNet} \citep{bernnet}: This model uses the Bernstein basis for approximation:
\begin{equation*}
    \hat{Y} = \sum_{k=0}^K \theta_k \frac{1}{2^K} \binom{K}{k} (2I - \hat{L})^{K-k} \hat{L}^k f_W(X)
\end{equation*}
{where $X$ is the input feature matrix, $\Theta=[\theta_0,\theta_1,\ldots, \theta_K]$ is the graph convolution parameter, $W$ is the feature transformation parameter and $f_W(X)$ is usually a 2-layer MLP. $\hat{L}$ is the normalized Laplacian matrix.

\section{Datasets}
The statistical information of the datasets, including node numbers, edge number, feature dimensions, node class numbers, edge homophilic ratios are summarized in in~\cref{table_dataset}.

We use the directed clean version of Chameleon and Squirrel provided by~\citep{critical-heter} which removes repeated nodes in graphs.
The large heterophilic dataset is proposed in~\citep{critical-heter}. The datasets Tolokers, Minesweeper and Questions are classified as homophilic datasets under the $H_{edge}$ metric~\citep{beyondh}, although they belong to heterophilic datasets according to the \emph{adjusted homophily} metric in~\citep{critical-heter}.

\section{Hyper-parameter Settings}
All experiments are run on a GPU NVIDIA RTX A6000 with 48G memory.

For all datasets, we set the maximum number of epochs to 1,000 and apply early stopping with a patience of 200 epochs. For the five large heterophilic datasets (Roman\_Empire, Amazon\_Ratings, Tolokers, Minesweeper, Questions), we fix the hidden size of the MLP to 512. For the remaining datasets, we set the hidden size to 64.

We conduct a grid search for hyper-parameters used during the training of spectral GNNs, including dropout rate, learning rate, exponential decay parameter, early stopping, and maximum training epochs. The exact search ranges for different hyper-parameters are detailed  in~\cref{table-asym-appendix-hyper}.

\begin{table}[ht]
    \centering
   \resizebox{1.0\columnwidth}{!}{%
    \begin{tabular}{c |c|c }
    \toprule
      Hyper Parameters & GNNs & Range\\
      \toprule
    dropout in MLP & All & $0.5,0.7,0.8,0.9$\\
    dropout after MLP & All & $0.5,0.7,0.8,0.9$\\
    learning rate of $\Theta$ &All & $0.001,0.01,0.05$\\
    weight decay of $\Theta$ & All & $0.0,0.0001,0.0005$\\
    weight decay of $W$ & All & $0.0,0.0001,0.0005$\\
    learning rate of $W$ & All & $0.005, 0.01, 0.05$\\
    $\beta_{\pi_{\Theta}}$ & All & $0.9,0.99$\\
    $\beta_{\pi_W}$ & All & $0.9,0.99$\\
    $a$ & JacobiConv & $-2.0,-1.0,-0.5,0.0,0.5,1.0,2.0$\\
    $b$ & JacobiConv & $-2.0,-1.0,-0.5,0.0,0.5,1.0,2.0$\\
    initialization parameter $\alpha$ & JacobiConv & {$0.1,0.5,0.9,1.0, 2.0$}\\
        initialization parameter $\alpha$& GPRGNN &{$0.1,0.5,0.9,1.0, 2.0$}\\
 \bottomrule
    \end{tabular}
   }
    \caption{Search ranges of hyper parameters.}
    \label{table-asym-appendix-hyper}
\end{table}

\begin{table*}[b]
    \centering
    \begin{tabular}{*{6}{c}}
        \toprule
        Dataset & Texas & Wisconsin & Cornell & Actor & Chameleon \\ 
        \midrule
        $\rho_{\Psi_1}$ 
        & $0.011089$ 
        & $0.016569$ 
        & $0.020607$ 
        & $0.009565$ 
        & $0.010492$ \\
        $\rho_{\Psi_2}$ 
        & $0.011085$ 
        & $0.016552$ 
        & $0.020638$ 
        & $0.009590$ 
        & $0.010541$ \\
        $\lambda_{max}(H_{\Psi_1})$ 
        & $0.4357$ 
        & $0.4470$ 
        & $0.5855$ 
        & $0.2255$ 
        & $4.3220$ \\
        $\lambda_{max}(H_{\Psi_2})$ 
        & $0.4347$ 
        & $0.4460$ 
        & $0.5859$ 
        & $0.2281$ 
        & $4.3205$ \\
        $\rho_{\Psi_1} \geq \rho_{\Psi_2}$ 
        & True 
        & True 
        & False 
        & False 
        & False \\
        $\lambda_{max}(H_{\Psi_1}) \geq \lambda_{max}(H_{\Psi_2})$ 
        & True 
        & True 
        & False 
        & False 
        & True \\
        \bottomrule
    \end{tabular}

    \begin{tabular}{*{5}{c}}
        \toprule
        Dataset & Squirrel & Citeseer & Pubmed & Cora \\ 
        \midrule
        $\rho_{\Psi_1}$ 
        & $0.013374$ 
        & $0.001995$ 
        & $0.005756$ 
        & $0.005261$ \\
        $\rho_{\Psi_2}$ 
        & $0.013391$ 
        & $0.001994$ 
        & $0.005721$ 
        & $0.005267$ \\
        $\lambda_{max}(H_{\Psi_1})$ 
        & $0.2683$ 
        & $0.1068$ 
        & $0.0933$ 
        & $0.1043$ \\
        $\lambda_{max}(H_{\Psi_2})$ 
        & $0.2687$ 
        & $0.1066$ 
        & $0.0931$ 
        & $0.1044$ \\
        $\rho_{\Psi_1} \geq \rho_{\Psi_2}$ 
        & False 
        & True 
        & True 
        & False \\
        $\lambda_{max}(H_{\Psi_1}) \geq \lambda_{max}(H_{\Psi_2})$ 
        & False 
        & True 
        & True 
        & False \\
        \bottomrule
    \end{tabular}
    \caption{Empirical data supporting~\cref{proposition-asym-speed-hessian}.}
    \label{table-asym-assump11}

    \resizebox{0.8\textwidth}{!}{
        \begin{tabular}{*{10}{c}}
    \toprule
       Iteration t   & 100 & 150 & 200 & 250 & 300 & 350 & 400 & 450 & 500 \\
       \toprule
       $\frac{s_{\Theta}^t}{s_W^t}$   &  1.5015 &  2.2624& 3.1904 &  2.8585 &  6.6780 &  4.1562 &      6.7149 &  11.5285 &  25.3993\\
       $\frac{\lambda_{max}(\bbH_{W,W}^t)}{\lambda_{max}(\bbH_{\Theta,\Theta}^t)}$ &  3.7220 &  0.8198 &  2.2311 &      2.1035 &  2.8941 &  2.6736 &  3.8299 & 11.0213 &  12.8264\\
       \bottomrule
    \end{tabular}
    }
    \caption{Empirical data supporting~\cref{assumption-asym-scale}.}
    \label{table-asym-assump13}
        \resizebox{0.9\textwidth}{!}{
   \begin{tabular}{*{10}{c}}
   \toprule
  Dataset & Texas & Wisconsin & Cornell & Actor & Chameleon & Squirrel  & Citeseer & Pubmed & Cora\\ 
  \toprule
  \(\lambda_{max}(H_{\Theta,\Theta})\) & 0.2681 & 1.0724 &1.5734 & 0.1954 & 2.4245 & 0.0596 & 0.0042 & 0.1711 & 0.1891\\
    \(\lambda_{max}(H_{W,W})\) & 0.0219 & 0.0249 & 0.0138 & 0.0182 & 0.0093  & 0.0252 & 0.0065 & 0.0165 & 0.0299 \\
  \(\lambda_{max}(H)\) & 0.4504 & 1.2951 & 1.6910 & 0.2134 & 2.4449 & 0.0890 & 0.0455 & 0.1723 & 0.2408\\
\bottomrule
\end{tabular}
}
\caption{Empirical data supporting underlying assumption that $\lambda_{max}(H)\geq \lambda_{max}(H_{\Theta,\Theta}) \geq \lambda_{max}(H_{W,W})$.}
\label{table-asym-underlying-assump}
\end{table*}

\section{Additional Experimental Results}
\cref{table-uniform-appendix-large-heter} shows the results of spectral GNNs of asymmetric learning on five large heterophilic graphs.
The ROC AUC metric is reported for Minesweeper, Tolokers, and Questions datasets as they contain only two node classes. In binary classification, AUC is preferred over accuracy because it better accounts for class imbalance.

We observe an average performance improvement ranging from $0.51\%$ to $1.74\%$ across different spectral GNNs, which demonstrates the effectiveness of asymmetric learning. The performance improvement on five large heterophilic datasets in~\cref{table-uniform-appendix-large-heter} is less significant compared to that on six small heterophilic datasets, as shown in~\cref{table-small-heter}. This can be attributed to the fact that with more training samples, the gradient estimates become more accurate. In other words, asymmetric learning provides great performance improvement, particularly when training samples are limited.

\section{Empirical Validation of Assumptions}
We conduct experiments to empirically validate the assumptions underpinning our theoretical analysis.

Specifically, empirical data supporting~\cref{proposition-asym-speed-hessian} is presented in~\cref{table-asym-assump11}. According to this assumption, for two points $\Psi_1$ and $\Psi_2$ near a critical point, if $\lambda_{\max}(\Psi_1) > \lambda_{\max}(\Psi_2)$, then $\rho_{\Psi_1} \geq \rho_{\Psi_2}$. Here, $\lambda_{\max}(\Psi)$ denotes the largest eigenvalue of the Hessian at point $\Psi$, and $\rho_{\Psi}$ represents the Gradient-Parameter Norm Ratio (GPNR). To test this, we select the point with the minimum empirical loss during training as the critical point. We then add random noise with a scale of $0.1$ to generate two nearby points. For these points, we compute their GPNR and the largest eigenvalue of the Hessian. On most datasets, this assumption is observed to hold, supporting the validity of our theoretical proposition.

Empirical data supporting~\cref{assumption-asym-scale} is shown in~\cref{table-asym-assump13}. We train ChebyNet on Texas using gradient descent and show the two ratios every 50 iterations. We observe that there is a high probability that 
\[
\frac{s_{\Theta}^t}{s_W^t} \geq \frac{\lambda_{\max}(\bbH_{W,W}^t)}{\lambda_{\max}(\bbH_{\Theta,\Theta}^t)}.
\]

For the underlying assumption 
\[
\lambda_{\max}(\bbH^t) \geq \lambda_{\max}(\bbH_{\Theta,\Theta}^t) \geq \lambda_{\max}(\bbH_{W,W}^t)
\]
in~\cref{theorem-precondition}, we show that this assumption holds on real-world graph datasets. Specifically, we present the values of $\lambda_{\max}(\bbH)$, $\lambda_{\max}(\bbH_{\Theta,\Theta})$ and $\lambda_{\max}(\bbH_{W,W})$ at the 50-th iteration when ChebyNet is trained on different datasets in~\cref{table-asym-underlying-assump}. We observe that the underlying assumption holds.

\end{document}